# Depth and Autonomy:
# A Framework for Evaluating LLM
# Applications in Social Science Research[*]

Ali Sanaei & Ali Rajabzadeh

`sanaei@uchicago.edu`, `rajabzadeh@methods.academy`

October 29, 2025

Large language models (LLMs) are increasingly utilized by researchers across a wide range of domains, and qualitative social science is no exception; however, this adoption faces persistent challenges, including interpretive bias, low reliability, and weak auditability. We introduce a framework that situates LLM usage along two dimensions, interpretive depth and autonomy, thereby offering a straightforward way to classify LLM applications in qualitative research and to derive practical design recommendations. We present the state of the literature with respect to these two dimensions, based on all published social science papers available on Web of Science that use LLMs as a tool and not strictly as the subject of study. Rather than granting models expansive freedom, our approach encourages researchers to decompose tasks into manageable segments, much as they would when delegating work to capable undergraduate research assistants. By maintaining low levels of autonomy and selectively increasing interpretive depth only where warranted and under supervision, one can plausibly reap the benefits of LLMs while preserving transparency and reliability.

# 1 Introduction

Large language models (LLMs) have been transformative for natural language processing and are increasingly used across qualitative social science research applications for indexing, summarization, first-pass coding, and more. Despite the excitement, however, adoption is constrained by concerns regarding interpretive bias, reliability, and auditability. Our goal is to propose a framework which helps us more easily classify, recommend, and evaluate the utilization of these models in research.

Our objectives are higher research quality (by which we mean validity, reliability, and interpretive coherence), greater transparency and reproducibility, and preserved human control.[1] We propose a two-dimensional plane along 'interpretive-depth' and 'autonomy,' and contend that the above objectives are jointly advanced by systematic constraints on model autonomy—by ensuring that LLMs operate as assistants without authority over consequential interpretive decisions—and allowing interpretive depth to vary according to the substantive goals of the analysis, under human supervision.

Our argument proceeds from a simple premise: contemporary models are powerful processors of natural language but remain brittle in settings that require hermeneutic inference, contextual sensitivity, expert knowledge, or reflexive judgment. We have acquired assistants that exceed human capabilities in some tasks (some examples are reviewed in Section 2) but have vexing deficiencies in other tasks. A growing body of evidence indicates persistent failures in complex comprehension, global reasoning, and narrative verification (Subbiah et al. 2024; Gevers et al. 2025; Manikantan et al. 2025; Hardt 2023b; Cui et al. 2023; Den-tella et al. 2024, 2024); They retain measurable social biases even when explicit tests appear neutral, with prompt-based diagnostics revealing implicit associations and discriminatory decision tendencies (Bai et al. 2025; Guo et al. 2024; Aguda et al. 2025). These models' shaky meta-cognition and lack of humility can only make the situation more precarious (Betley et al. 2025; Anthropic 2025). From a teleological perspective, many of these behaviors correspond to the statistical objectives and data distributions that shape next-token prediction, leaving detectable "embers of autoregression" in model behavior even as capabilities improve and even in models optimized for reasoning (R. Thomas McCoy et al. 2023; R Thomas McCoy et al. 2024), and the question of 'how high the LLM asymptote is' does not have an empirical or theoretical answer yet.

---

1. A research project, from start to finish, can be partitioned into elements which should be reproducible and elements for which reproducibility is not important or maybe not possible. For example, how exactly one identifies a puzzle in a literature, or thinks of a theory, or devises a hypothesis need not be reproducible. The elements for which reproducibility is needed are the concern of the present paper. Of course, one can chat with a language model to come up with new hypotheses—that's outside our scope.

While model scaling, instruction tuning, reinforcement learning from human feedback, and rationale scaffolding have improved task following and multi-step reasoning, these advances are uneven and frequently sensitive to task form and evaluation design. Instruction-tuned systems demonstrate notable zero- and few-shot gains (Chung et al. 2022; Brown et al. 2020; Chowdhery et al. 2023; Ouyang et al. 2022), and chain-of-thought prompting can elicit performance improvements on arithmetic and commonsense benchmarks (Wei, Wang, et al. 2022; Wei, Tay, et al. 2022). The hallucination (and limited context) issues have been considerably remedied by using retrieval-augmented generation, which has now become standard practice for grounding outputs (Lewis et al. 2020; Izacard et al. 2023; Borgeaud et al. 2022; Huang and Huang 2024; Karimzadeh and Sanaei 2025).

Notwithstanding this success, replication studies indicate that many our present solutions are fragile across models and benchmarks, underscoring the need for standardized protocols, multiple seeds, and transparent documentation (Vaugrante, Niepert, and Hagendorff 2024). The totality of these observations imply that we have enticingly cheap and powerful tools at our disposals, but we must be cautious about what tasks are given to them, and having ways of preserving human control and supervision.

In response to this challenge, we adopt the bounded-autonomy principle: models may propose candidates, summarize evidence, and surface contrasts, but they should be prevented from making critical decisions or executing complex tasks without a clear roadmap. We contend that autonomy becomes more important as our tasks require higher levels of interpretive depth. Operationally, we constrain the LLM to research assistant roles with a clear rubric, worked examples, and tightly scoped subtasks; it must cite the textual basis of its suggestions, indicate uncertainty, and escalate difficult judgments to the human analyst. The human retains prerogatives over coding decisions, category formation, conflict resolution, and theoretical integration, much as a PI retains responsibility for research claims developed with the help of research assistants. This position is consistent with emerging practice in qualitative workflows that utilize LLMs as bounded aids for first-pass coding, code suggestion, and memo drafting while maintaining auditable artifacts (Dai, Xiong, and Ku 2023; Chew et al. 2023; Dunivin 2024; Sinha et al. 2024), and it aligns with frameworks that emphasize LLMs as tools to propose or refute models under direct human checking (Eschrich and Sterman 2024) and to structure multi-agent proposer–critic–adjudicator roles with logged exchanges (Rasheed et al. 2025; Su et al. 2024).

In the pages that follow, first, we formalize a depth by autonomy framework in Section 2 that yields design rules and evaluation criteria and show how vertical and horizontal decomposition can attain high interpretive depth under low autonomy through staged and auditable pipelines. Then, in Section 3 we presents the coding instruments and apply them to the existing literature to assess how the present literature can be projected on these two dimensions; We finally further empirical demonstrations. Finally, Section 5 concludes the paper, and replication materials appear in the Appendix.

## 2 Oracles or Bounded Assistants

The LLMs lack a conceptual notion of incapacity as they have been trained on internet-sized data, and have been trained to be all-knowing helpful assistants, which as a result encourage users to treat them as oracles. While that may be far-fetched, a cursory search for how LLMs are used in research, especially research that is not published yet, yield ample evidence of the naive optimism with which some researchers are relying on these models. In qualitative inquiry and as a matter of design and accountability, we posit that generative LLMs should be cast as assistants without agency over consequential interpretive moves; see (Roberts, Baker, and Andrew 2024; Schroeder et al. 2025). A practical heuristic is to treat the model as a competent student assistant in their sophomore year: provide a rubric and examples, require citations, and reserve authoritative decisions to the researcher. The model operates as a bounded tool for indexing, summarizing, labeling, proposing alternatives, or even deep-diving into a corpus to extract novel hermeneutic insights while humans retain decision prerogatives over the exact procedures and oversee the execution step-by-step.

A distinction in qualitative methodology separates surface-level descriptive tasks from deeper, more complex interpretive work. This distinction separates what Corbin and Strauss (2015) term 'superficial analysis' which 'skims the top of data,' from 'in-depth analysis' that 'digs beneath the surface […] to explore all possible meanings' (p. 86). Also, there is a distinction in content analysis method between quantitative and qualitative content analysis; While early content analysis focused on the 'objective, systematic and quantitative description of the manifest content of communication' (Berelson 1952, p. 18), qualitative approaches emphasize discovering meaning within texts through interpretive and hermeneutic engagement (Kracauer 1952). Some qualitative scholars have distinguished between "thick" and "thin" description. A key aspect of deeper analysis is the transition from thin description—merely stating facts—to thick description, which includes the context, intentions, and meanings that underlie an action (Dey 1993; Denzin and Lincoln 2017). As Kuckartz indicates, meaning is often unreachable without prior knowledge, as understanding a text requires context that cannot be inferred from the text independently and cannot be automated or isolated into discrete parts; he has also recognized a correlation between knowledge and the ability to identify layers of meaning, suggesting that the more someone knows, the more levels of meaning they can understand (Kuckartz, 2014). Methodological frameworks, such as grounded theory, are explicitly designed to facilitate this advancement from description toward theoretical construction. This is achieved through a phased analytical process that begins by "fracturing the data" in initial open coding before moving to abstract conceptualization (Tie, Birks, and Francis 2019; Corbin and Strauss 2014; Creswell and Creswell 2022; Denzin and Lincoln 2017). Subsequent stages, such as "axial coding", systematically reconnect these concepts by examining their relationships through a paradigm of conditions, context, actions, and consequences, culminating in "selective coding", where a core category is identified and integrated with other categories to form a coherent theoretical account. This analytical climb involves moving from basic-level concepts, which are close to the raw data, to higher-level, more abstract categories that capture a central theme or phenomenon. Achieving this level of abstraction is not a mechanical task but relies on interpretive techniques that require human judgment, such as constant comparison, analyzing metaphors and emotional expressions, and maintaining the analytical distance needed to "walk a fine line between getting into the hearts and minds of respondents while at the same time keeping enough distance to be able to think clearly and analytically" (Corbin and Strauss 2014). Ultimately, in-depth qualitative work depends on the researcher's accumulated knowledge—Recognizing, as Dey (1993) puts it, that there is "a difference between an open mind and an empty head"—to transform descriptive data into a conceptual or theoretical contribution(Dey 1993).

Figure 1 presents Tesch's taxonomy of qualitative research, where she has organized methods according to whether they target the characteristics of language, the discovery of regularities, or the comprehension of meaning (1990). As one moves down the taxonomy, the analysis becomes increasingly concerned with latent meaning, theoretical embedding, and hermeneutic interpretation. This gradient is central for our purposes, since it implies that the extent to which a task depends on latent constructs and hidden context is likely to correlate inversely with model reliability when autonomy is high.

Recent evaluations of LLM-assisted qualitative tasks, for example, report strong performance on content extraction and shallow categorization but mixed results on tasks requiring context integration or interpretive synthesis (Bojic et al. 2025; Heseltine and Clemm von Hohenberg 2024; Friedman, Owen, and VanPuymbrouck 2024); complementary mappings and interview studies document similar tensions in adoption and evaluation (Schroeder et al. 2025; Barros et al. 2025).

On the capability side, progress is rapid, and LLMs have surpassed human capabilities in solving boutique linguistic tasks like multiple center embedding and garden path sentences, and have gained emergent human capabilities like theory of mind (Hardt 2025; Kosinski 2024). It is difficult to imagine a human who could read the following sentence easily:

> The cheese that the mouse that the cat that the dog that the boy that the teacher that the principal that the inspector noted reported warned scolded chased caught ate was moldy.

This was generated (and understood) by gpt-5, and even an open-weight model like `qwen3-32b` had no problem resolving it even without 'reasoning'.[2] But there is a different side to the story: LLMs routinely struggle with deeper levels of meaning in summarizing never-seen-before texts (Subbiah et al. 2024), in resolving ellipsis (Hardt 2023a), in book-length claim verification (Karpinska et al. 2024) and they misrepresent sources and context in real-world

---

2. The prompt was `Turn this sentence into simple short sentences: `The cheese that the mouse that the cat that the dog that the boy that the teacher that the principal that the inspector noted reported warned scolded chased caught ate was moldy.'`

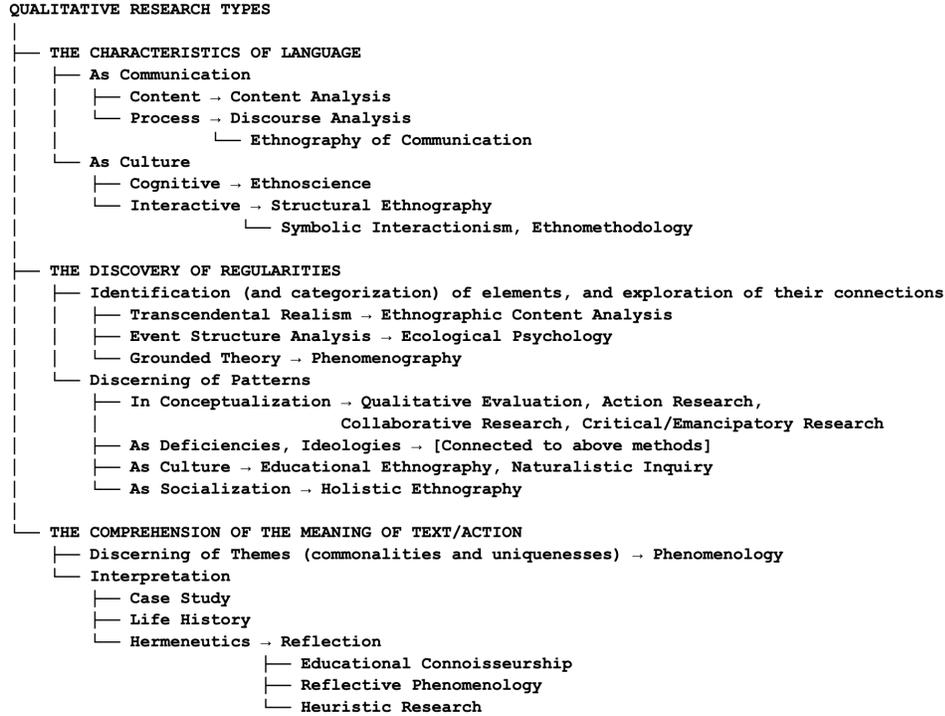

```
QUALITATIVE RESEARCH TYPES
|
├── THE CHARACTERISTICS OF LANGUAGE
|   ├── As Communication
|   |   ├── Content → Content Analysis
|   |   ├── Process → Discourse Analysis
|   |   |           └── Ethnography of Communication
|   └── As Culture
|       ├── Cognitive → Ethnoscience
|       └── Interactive → Structural Ethnography
|                         └── Symbolic Interactionism, Ethnomethodology
|
├── THE DISCOVERY OF REGULARITIES
|   ├── Identification (and categorization) of elements, and exploration of their connections
|   |   ├── Transcendental Realism → Ethnographic Content Analysis
|   |   ├── Event Structure Analysis → Ecological Psychology
|   |   └── Grounded Theory → Phenomenography
|   └── Discerning of Patterns
|       ├── In Conceptualization → Qualitative Evaluation, Action Research,
|       |                          Collaborative Research, Critical/Emancipatory Research
|       ├── As Deficiencies, Ideologies → [Connected to above methods]
|       ├── As Culture → Educational Ethnography, Naturalistic Inquiry
|       └── As Socialization → Holistic Ethnography
|
└── THE COMPREHENSION OF THE MEANING OF TEXT/ACTION
    ├── Discerning of Themes (commonalities and uniquenesses) → Phenomenology
    └── Interpretation
        ├── Case Study
        ├── Life History
        └── Hermeneutics → Reflection
                          ├── Educational Connoisseurship
                          ├── Reflective Phenomenology
                          └── Heuristic Research
```

Figure 1: Tesch's taxonomy of qualitative research types.

news answering (Archer and Elliott 2025). Moreover, this all happens with high levels of confidence, and lack of meta-cognition (Chen et al. 2025). They also have an instruction-following problem: they may assume more liberties than they are given, or they may be lazy in performing multi-step tasks (Lou, Zhang, and Yin 2024; Zhao et al. 2024; Hernández-Orallo et al. 2024; Tang et al. 2023).

There are two main strategies at play to try to resolve this tension between super-human power and second-hand Dunning-Kruger-esque combination of confidence and incompetence: first, the technical aspects, which is progressing with full-speed and is reducing error and bias either by providing better models, or by introducing remedies like 'reasoning' and 'grounding facts with web search,' but are out of the hands of most social scientists; and, second, by coming up with better research designs, that help rely on these models, for their strengths, avoid their weaknesses, and produce reliable results. This is where our focus lies; By propos-ing a framework for comparing various applications of LLMs in different fields, we hope to help establish better research designs and develop a language to evaluate research designs.

## 2.1 Dimensions of LLM Usage

Let us begin by delineating several potential dimensions along which LLM usage in qualitative text analysis can be characterized.

- Depth of analysis (surface ↔ hermeneutic) denotes the extent to which outputs rely on manifest linguistic features versus latent thematic or interpretive inference.

- Autonomy level (from tool-like "assistant" to "delegate" to "trustee") denotes the extent to which consequential choices are made by the model rather than by a human.

- Scope of analysis (going from word to sentence to segment to document to corpus) refers to the unit of analysis on the input side and by the nature of the task.

- Reasoning load (simple recall ↔ multi-step reasoning) indexes whether performance is plausibly pattern retrieval or requires explicit multi-step inference. Like other dimensions, this is about the task, not the model. For example, on the easier side of the spectrum, imagine going from 'What state contains Albuquerque?' to 'Name all states that start with the same letter that the name of the state that contains Albuquerque starts with but does not contain Albuquerque.'

- Task novelty (in-training ↔ novel) distinguishes prompts that resemble training patterns from genuinely new problems. In the former case, models typically perform well irrespective of the task's complexity. In the latter case, model performance relies on how well can the existing training come to the rescue (one might say, like how humans perform new tasks), even if the model has not 'performed' the task in training, it may have seen it, like answering medical diagnostic questions.

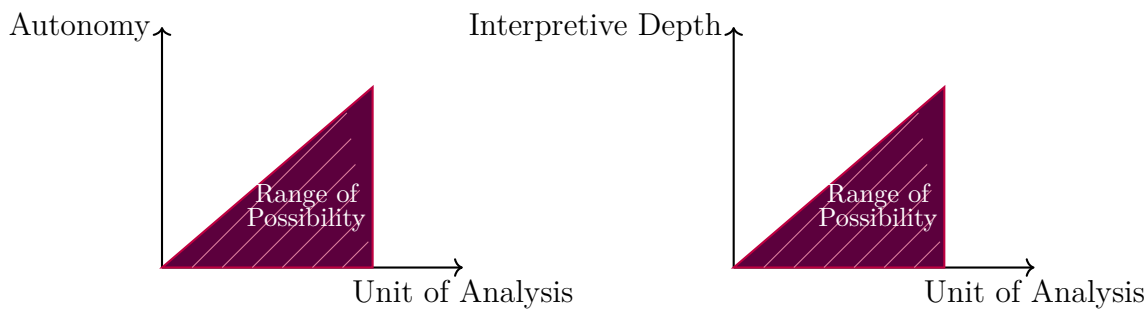

Figure 2: Research methodology constraint plots showing feasible regions

- Inference (descriptive ↔ interpretive) specifies whether the task summarizes observable content or imputes latent constructs.

- Logic (deductive ↔ inductive) encodes whether categories are fixed a priori or emerge iteratively.

- Context (contextual ↔ non-contextual) indicates the extent to which broader situational information must be integrated.

- Iteration (iterative ↔ single-shot) captures whether the pipeline is multi-pass or single-pass, including multi-agent variants (Rasheed et al. 2025).

- Epistemology (positivist ↔ interpretivist) situates the epistemic stance of the analysis (Eschrich and Sterman 2024).

These dimensions exhibit systematic correlations. For example, figure 2 demonstrates the constraint relationships between scope and both autonomy and interpretive depth; the feasible regions expand with analytical scale, revealing how broader context enables—but does not necessitate—greater model agency or hermeneutic complexity. While these dimensions overlap, we argue that interpretive depth and autonomy are conceptually distinct and directly actionable in design. Depth is set by the substantive aim of the study; autonomy is set by the pipeline. Together, they define, for a given task, what the model is able to do and what must be reserved for humans. Moreover, these two axes subsume, in terms of predictive leverage, a wide range of the other dimensions—scope, novelty, and reasoning load, inference, logic, context, iteration, and epistemology—are all easy to relate to these two dimensions, in an abstract way, although the exact relationships depend on the specific task and the context.

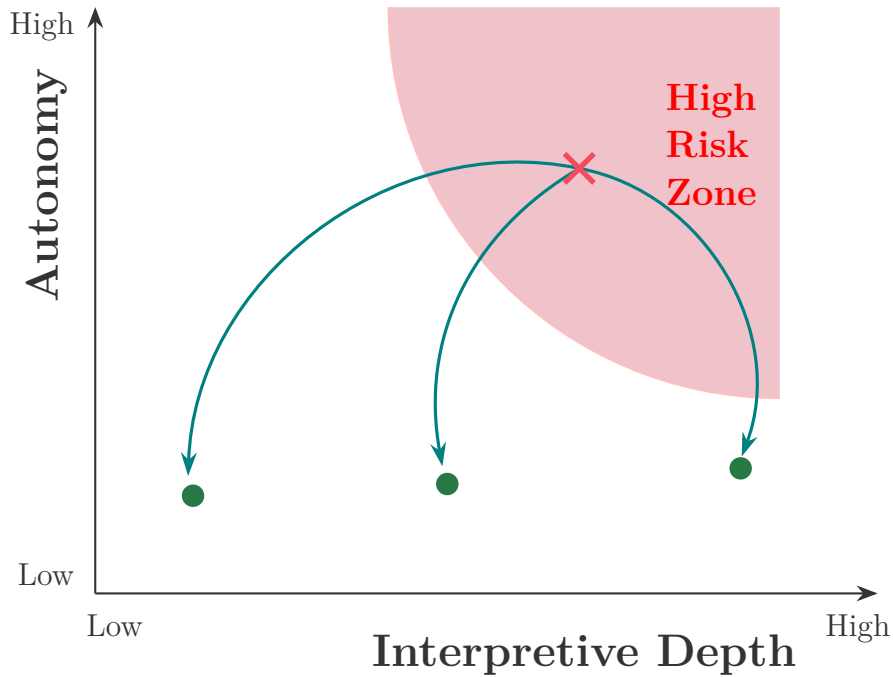

Figure 3: Depth and autonomy: configurations and risk region. Low-autonomy configurations (green points) can support increasing interpretive depth; the shaded sector marks high-risk high-autonomy/high-depth configurations.

It deserves emphasis that the interpretive depth associated with a substantive research question is distinct from the depth of the operation assigned to the model. The latter is a function of the protocol: What exactly is the model asked to do? What examples are supplied? Which outputs are permitted?

## 2.2 Context, Depth, and Autonomy

As Figure 2 illustrated, the feasible set for what the model could do expands with context. The methodological problem is that realized autonomy can expand in lockstep with this feasible set when researchers delegate end-to-end tasks to a model. Our advice is to try to break this coupling by design when possible: a concern reinforced by observed directional biases in relation predictions and implicit associations (Aguda et al. 2025; Bai et al. 2025; Guo et al. 2024). We allow interpretive depth to rise when warranted by the research question, but we leash realized autonomy through bounded subtasks, structured outputs, and mandatory human checkpoints.

A practical corollary concerns task decomposition. Two strategies are useful in this setting. Vertical decomposition sequences subtasks so that the input to stage $k$+1 is the output of stage $k$ (e.g., extract evidence → cluster codes → synthesize themes). Horizontal decomposition, in contrast, runs tasks in parallel—either across disjoint input segments when context budgets are binding (chunking) or across distinct dimensions applied to the same input (e.g., rule of law, accountability, institutional constraints). Earlier models often required horizontal decomposition because they drifted when requested to perform multiple tasks concurrently and faced context limitations on long texts. While contemporary systems are more capable, task decomposition typically produces richer outputs, more faithful instruction following, and multiple checkpoints that improve transparency and autonomy control by creating opportunities to diagnose and correct intermediate artifacts.

## 2.3 Orchestrated Decomposition on the Autonomy-Depth Plane

Much can be accomplished by research design. Most importantly, high interpretive depth does not necessitate high autonomy, as it may be possible to decompose the workflow ('bound' it), make it auditable, and have steps that require human approval. Single-pass execution concentrates latent decisions in one opaque step. In contrast, multi-pass pipelines separate extraction, candidate generation, adversarial critique, and adjudication, thereby distributing depth across stages while maintaining low autonomy at each stage. In this configuration, depth increases through synthesis across documented steps, rather than through early delegation to a model.

Design rule: When interpretive depth is high or the stakes of inference are substantial, utilize vertical decomposition to separate decision-bearing steps and horizontal decomposition to diversify inputs or dimensions. Each stage should have a narrow brief, typed outputs, calibrated abstention, and a documented handoff. The objective is to preserve low realized autonomy throughout, while enabling richer interpretive synthesis at the end of the pipeline.

The following three items are presented as examples, not prescriptions. Each demonstrates how high- or moderate-depth interpretive work can be implemented under low autonomy, using staged, auditable designs. Of course, all LLM steps can be iterated (to arrive at a satisfactory prompt) and can be run multiple times (to have a better sense of the uncertainty from the model's side).

*Example 1.* Extracting elements of constitutional thought from a 7th century document. The document is a letter from Ali ibn AbiTalib (the second caliph for Sunni muslims and the first imam of Shia muslims) to Malik al-Ashtar, his governor for Egypt in year AD 659. This is, while not the deepest task (especially given the roughly 3000 word length of the document), still requires significant interpretive depth.

Our decomposition plan is as follows: (i) Extract dimensions of constitutional thought from the sources, with clear definitions and evidence expectations; (ii) Run the model on the document, for each dimension, to provide a short explanation of whether that dimension is absent or present in the document, and if it is present, provide direct verbatim quotations that support the claim. (iii) Adjudicate between different claims. (iv) Synthesize the results into a final report.

*Example 2.* Open coding of archival radio transcripts. Our decomposition plan is as follows: (i) elicit candidate descriptive codes on short segments using worked examples and descriptions of what is intended, and inclusion of an abstention option; (ii) human consolidation into a provisional codebook; (iii) parallel application with abstention and conflict flags; (iv) adversarial pass proposing merges, splits, and negative cases with citations; (v) human revision; (vi) full-corpus application with reconciliation. Depth is moderate in the synthesis phases; autonomy remains bounded by the rubric, abstention, and human adjudication.

*Example 3.* Focus-group synthesis for marketing insight. We can imagine a pipline like this:(i) extract claims, needs, and quotations with source linkage (low depth); (ii) cluster (maybe by persona, maybe by general stance, etc.) (moderate depth); (iii) produce evidence-linked opportunity statements with confidence ratings and counter-evidence; (iv) human prioritization; (v) recommendation drafting with traceable links back to evidence and explicit caveats. The model proposes options and clarifies trade-offs; humans decide priorities and finalize language.

We can generalize the idea beyond these cases by mapping qualitative method families to the autonomy-depth plane in order to derive role assignments for models and humans. The idea can be summarized in this pithy slogan: *Break the task, bind the output, and climb the ladder of abstraction under human gaze.*

# 3   Survey of LLM Use in Social Science Research

In the preceding section we claimed that the depth-autonomy framework can both guide our design decisions and can also help us evaluate existing research. Here we develop a coding scheme that we apply to existing published social science research that has utilized generative LLMs.

Table 1: Summary of items in the coding questionnaire; full instrument in Appendix A

| Construct | Items (abridged descriptions; scoring anchors) |
| --- | --- |
| Description | Q01-Q09 (discipline, data type/language, …) |
| Interpretive depth | Q10 Task nature (1-5: extraction…deep interpretation); |
| | Q11 Ambiguity (0-2); |
| | Q12 External context (0-3); |
| | Q13 Reasoning (0-4); |
| | Q14 Framework predefined vs emergent (1-3); |
| | Q15 Unit of analysis (1-5) |
| Realized autonomy | Q16 Human scaffolding (0-3); |
| | Q17 Human supervision (0-3); |
| | Q18 Instruction mode (interactive/fixed/agentic); |
| | Q20-Q21 Reasoning prompts and examples (yes/no); |
| Transparency, Validation | Q23-Q33 (model identification, prompts shared, evaluation against humans, limitations) |

## 3.1 Measurement model and coding instrument: overview

Our instrument operationalizes two central constructs: interpretive depth (what kind of inference the model is tasked to perform) and realized autonomy (the extent to which consequential steps are delegated to the model versus scaffolded and supervised), but also includes items to collect descriptive metadata (discipline, data types, language), other aspects of the design (unit of analysis), evaluation practices (validation against humans, reporting of limitations), and transparency of research (whether and to what extent replication materials are shared). The full instrument, coder instructions, tie-breakers for primary-use selection, and the rationale-and-evidence protocol appear in Appendix A.

Table 1 summarizes the mapping between constructs and items. Items Q10-Q15 index interpretive depth. Items Q16-Q22 index realized autonomy. Items Q01-Q09 and Q23-Q33 are descriptive and evaluation covariates. The item content and anchors align with the conceptual framework developed in …

**Exemplar items.** (Full text in Appendix A)

Q10-Nature of the task performed by the LLM. (1) Information extraction; (2)

Summarization/synthesis of explicit content; (3) Initial qualitative coding (surface); (4) Thematic analysis (latent); (5) Deep interpretation (theory-building).

Q16-Human scaffolding of the task (end-to-end pipeline). (0) Not decomposed; (1) Small extent; (2) Moderate extent; (3) Large extent (detailed checklist/codebook).

Q17-Human supervision of the LLM's work. (0) None; (1) Occasional; (2) Regular; (3) Intensive (approval at each step).

## 3.2  Empirical Results

We queried the Web of Science Core Collection to identify social-science journal articles that deploy large language models (LLMs) in substantive research.[3] The query returned 955 records. Five lacked abstracts, leaving 950 for screening. We then conducted a three-pass screening protocol. First, using DeepSeek-Reasoner v3.1 (temperature = 0), we classified abstracts as relevant if an LLM was used instrumentally—e.g., as an analytic tool, coding assistant, data generator under constraints, or research aide in an actual study—rather than if the paper's sole object was to analyze or benchmark LLMs themselves. This stage yielded

---

3. We queried the Web of Science Core Collection on 26 August 2025 at 16:30 UTC to identify social-science journal articles that deploy large language models (LLMs) in substantive research. The search expression was: TS=(large language model OR LLM OR GPT) AND PY=(2023–2025) AND DT=(ARTICLE OR EARLY ACCESS) AND WC=(Social Sciences, Interdisciplinary; Communication; Behavioral Sciences; Law; Social Sciences, Mathematical Methods; Political Science; Psychology, Social; Psychology, Multidisciplinary; Social Issues; Sociology; History; Anthropology; Religion; Social Work; International Relations). We included GPT" because many authors name that system explicitly in titles and abstracts after the release of ChatGPT, whereas "LLM" is used inconsistently across fields. The 2023–2025 window captures the period when generative models entered applied workflows while accommodating indexing lag. Restricting results to Article and Early Access concentrates the output on peer-reviewed journal material and The disciplinary filter spans political science and adjacent social-science fields to ensure coverage across cognate domains.

234 items. Second, we re-screened these abstracts twice with GPT-5 (reasoning effort set high) and retained items judged relevant across all three LLM calls. Third, we manually adjudicated the resulting set, removing papers that did not actually use generative models, used them only as the object of study, or offered comparisons of models without instrumental use. The final corpus contains 56 articles; we also retrieved their PDFs. Our aim was not exhaustiveness. We sought a diverse corpus of social-science studies that actually employ LLMs in empirical or analytical work.

We applied the coding scheme to the retrieved works in three ways: first, we applied the coding scheme using a competent open-weight model (`gpt-oss-120b` with high reasoning) giving the codebook and the text of the papers, five times per paper, and asking all questions that needed some reasoning to also produce a clear rationale. A random review of the results proved disappointing with various types of mistakes: the majority of mistakes were those could be easily done by human assistants who do not pay close attention to details (multiple uses of generative models was one of the causes of some mixups); other mistakes were mistakes in degree, in how Likert-type questions in the codes were answered; but there was a third category of mistakes that were a bit baffling and interestingly all runs of the model would agree on their wrong answer, but could be corrected with few-shot examples. An example is when clear examples of 'classification' would be categorized as 'information extraction.'

```
<cot>
The task is to classify tweets into predefined issue categories (e.g.,
health, economy), which involves identifying explicit topics mentioned
in the text. This is information extraction, not summarization, coding,
or deeper interpretation.
</cot>
1: Information extraction (identify explicit facts)
```

We then used 'gpt-5' with high reasoning on the same data. The results were generally better, but residual confusions persisted: degree errors on Likert anchors and stable misclassifications that required few-shot guidance (as in the classification vs information-extraction example).

## 3.3 Construct-level variation

How does the literature look throught the lens of our framework? We present a short analysis that evaluates feasibility and variation for the autonomy–depth framework using the coded corpus. The analysis demonstrates that the questionnaire items in Table 1 can be operationalized with published materials, that the items are answerable with sufficient fidelity to construct definitions, and that the resulting indices exhibit non-trivial dispersion across studies. The objective is validation of implementability in the corpus, not hypothesis testing about structural relations among the constructs.

The measurement follows the coding instrument. Interpretive depth aggregates Q10 to Q15, which capture task nature, ambiguity, external context, reasoning, framework status, and unit of analysis, and Realized autonomy aggregates Q16 to Q17 and Q22, which capture human scaffolding, human supervision, and iteration. The instruction mode from Q18 is recorded at the item level and contributes to the autonomy item set as implemented. Reproducibility-and-rigor aggregates transparency and evaluation indicators: model identification, settings reporting, prompt availability, materials sharing across prompts, code, and data, evaluation against a human standard or benchmark, limitations discussion, and reliability reporting (Q23, Q25, Q27, the multi-item materials count, Q30–Q33). All items are rescaled to the unit interval prior to row-wise averaging with available cases. This available-case approach is intended to preserve information while avoiding listwise deletion. The indices are descriptive summaries rather than latent-variable estimates.

The analysis yields three indices with visible dispersion on the unit interval. Figure 4 summarizes marginal distributions and bivariate relationships. Pairwise correlations are re-ported strictly as descriptive markers of separability. Interpretive depth with autonomy equals $-0.14$, interpretive depth with reproducibility and rigor equals $-0.23$, and autonomy with reproducibility and rigor equals $-0.03$, computed with pairwise deletion. The magnitudes are modest, which is consistent with the intended use of these indices as distinct descriptive dimensions. The central result here is variation. The literature contains studies at different points in the autonomy–depth plane and with heterogeneous transparency and evaluation practices. This is what is needed for subsequent descriptive comparisons that utilize these indices as classification variables.

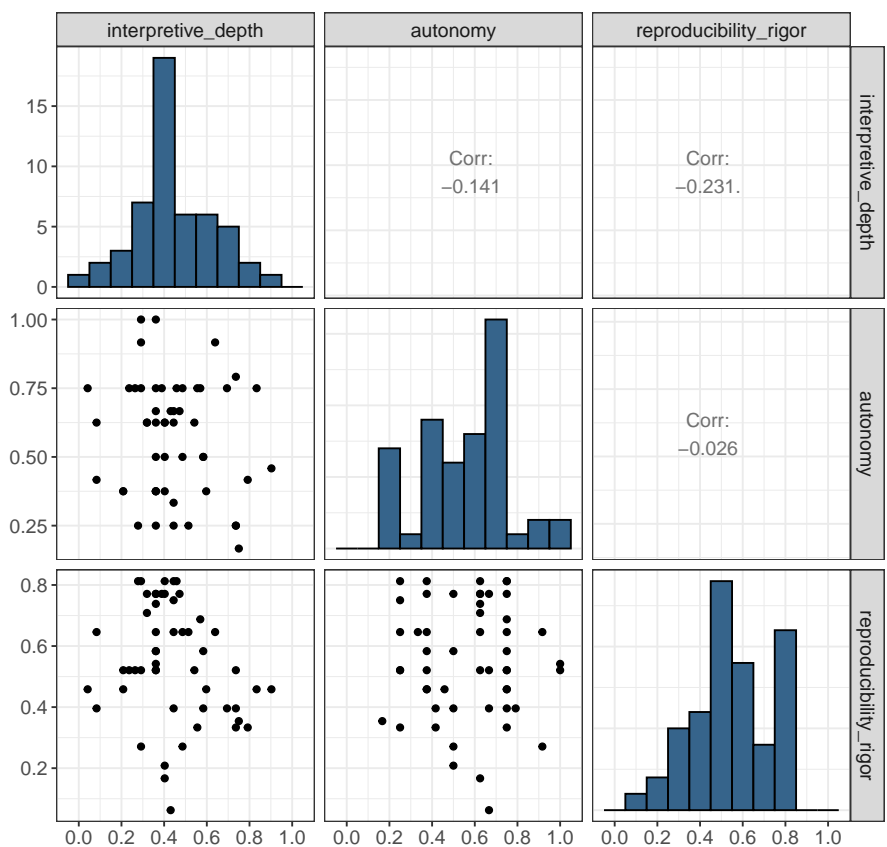

Figure 4: Correlations and distributions of the constructs in the literature corpus. The figure shows a scatterplot matrix for interpretive depth, realized autonomy, and reproducibility, each scaled to the unit interval. Off-diagonal panels show pairwise relationships. Diagonal panels show marginal distributions.

# 4 Results

In this section we report two empirical demonstrations designed to evaluate the bounded-autonomy principle. Implementation details, prompts, and full audit trails appear in Appendix A and Appendix B. Both of the experiments are tasks about "Letter 53" which is an edict by Imam ᶜAlī to Malik al-Ashtar (AD 659), which is a canonical governance directive when Malik was appointed governor of Egypt (al-Sharif al-Radhi 1987). All of the LLMs used here have had this letter and multiple translations of it in their training, but the tasks we are asking them to perform are novel and so the models training data do not comprise anything directly answering our specific tasks. Experiment 1 is an anachronistic and impossible task to demonstrate what could go wrong in the absence of guardrails and off-ramps; experiment 2 is about a legitimate theoretical question that seeks to evaluate this old text through the lens of modern ideas about governance.

## 4.1 Experiment 1: Prompt-Bounded Abstention on a Conceptually Mismatched Task

The goal is to assess how overly-compliant behavior by LLMs can lead to behavior that defies the user's intentions. The test case is deliberately asking to find for evidence that is so obviously absent, but all models we tested easily complied and provided some pieces of evidence with twisted arguments for why the irrelevant pieces could be seen as relevant.

The conclusion we want to draw is a strong word of caution: by reducing autonomy we do not mean limiting the choices of an LLM; rather we mean the freedom given the model to make *consequential decisions.* In the language of codebook development, a clearer codebook reduces coding errors by research assistants.

**Design**

The task is to "produce evidence of advocating for bicameralism" in a 7th-century piece of political advice (letter 53 of Nahjulbalaghah). We implement a $2 \times 2$ design with these factors: Enumerative range 0–10, 1–10 and Abstention option present, absent, with 50 runs per cell. In the control condition, the model (gpt-5; reasoning effort = medium; verbosity = medium) is instructed to extract "evidence elements" and return each item within an <evidence> tag. In the treatment condition, the identical prompt additionally states: "Or, you can say: 'There is no evidence for that!'" For each of the four cells we run 50 parallel calls on the same input letter. The primary outcome is the count of `<evidence>` tags per response, which, by construction, lies in [0,10]. Content validity is not adjudicated here because the task has no true positives; the correct output is abstention. All of the items we saw were utterly irrelevant, as expected, and with various twists in logic they were pushed as evidence supporting bicameralism. What was more informative was that the thinking provided by some models showed that the models clearly had a sense that the task was impossible or anachronistic, but still obsequiously complied, even when 0 was an option!

**Outcomes**

The quantity of enumerated "evidence elements" is used as a behavioral indicator of compliance versus abstention in this case. We report the sample mean and standard deviation of counts across the 50 runs per cell; we also record whether any run produced zero items.

Table 2 reports the distributional summaries. Without an abstention option, the model reliably fabricates between five and eight "evidence" items, depending on whether the enumerative range is 0-10 or 1-10. With an explicit abstention option, outputs collapse to zero almost always, including in the 1-10 setting where zero is not within the numeric range. For [0-10, no abstention], the mean count is 5.26 (SD= 1.85) and for [1-10, no abstention], it is 7.36 (SD= 0.964). Adding the explicit abstention string yields [0-10, abstention] mean

= 0.00 (SD= 0.00; 50/50 zero-count runs) and [1-10, abstention] mean = 0.16 (SD= 1.13), with 49/50 zero-count runs and one outlier run returning eight items.

Figure 5 shows the results of the experiment done with 4 top models. While there is some difference between the models, they all suffer from this behavior.

**Interpretation**

Two implications follow. First, in the case of an impossible task, hard enumerative bounds (e.g., "give 1-10 items") act as constraints that the model prioritizes satisfying, yielding nonsensical outputs rather than calibrated abstention. Second, "reduced autonomy" must include an explicit valid off-ramp, an abstention clause that is semantically consistent with the decision space, if we wish to prevent spurious compliance. Put differently, instructing a model to "stay within tight bounds" without an auditable abstention path risks reliability loss through over-compliance; adding a clear abstention option re-routes behavior toward refusal, even overriding numeric bounds in nearly all runs. The extent to which abstention is realized is therefore a function of prompt semantics as well as the allowed output set, and careful human supervision remains necessary to iterate prompts or to halt tasks that are ill-posed.

Recent research demonstrates the butterfly effect of changing minor characteristics such as spacing, punctuation, and adverbs (Salinas and Morstatter 2024; Sclar et al. 2024), changing the prompt structure (He et al. 2024; Salinas and Morstatter 2024), the order of instructions (e.g., reasoning first then scoring, or scoring followed by reasoning)(Chu, Chen, and Nakayama 2024), and semantically similar prompts (rephrasing prompt, changing language) (Barrie, Palaiologou, and Törnberg 2025; Errica et al. 2025; Stewart et al. 2024) could lead to significantly different outputs. While recent models have shown better consistency, neither model size nor prompt optimization methods, nor the use of reasoning models, has fully addressed this challenge (He et al. 2024; Sclar et al. 2024). There is a substantial need to refine the prompt based on the model and specific tasks, which means we require an iterative approach with human supervision checkpoints to minimize output inconsistency and improve output quality, while also enhancing replicability and reliability.

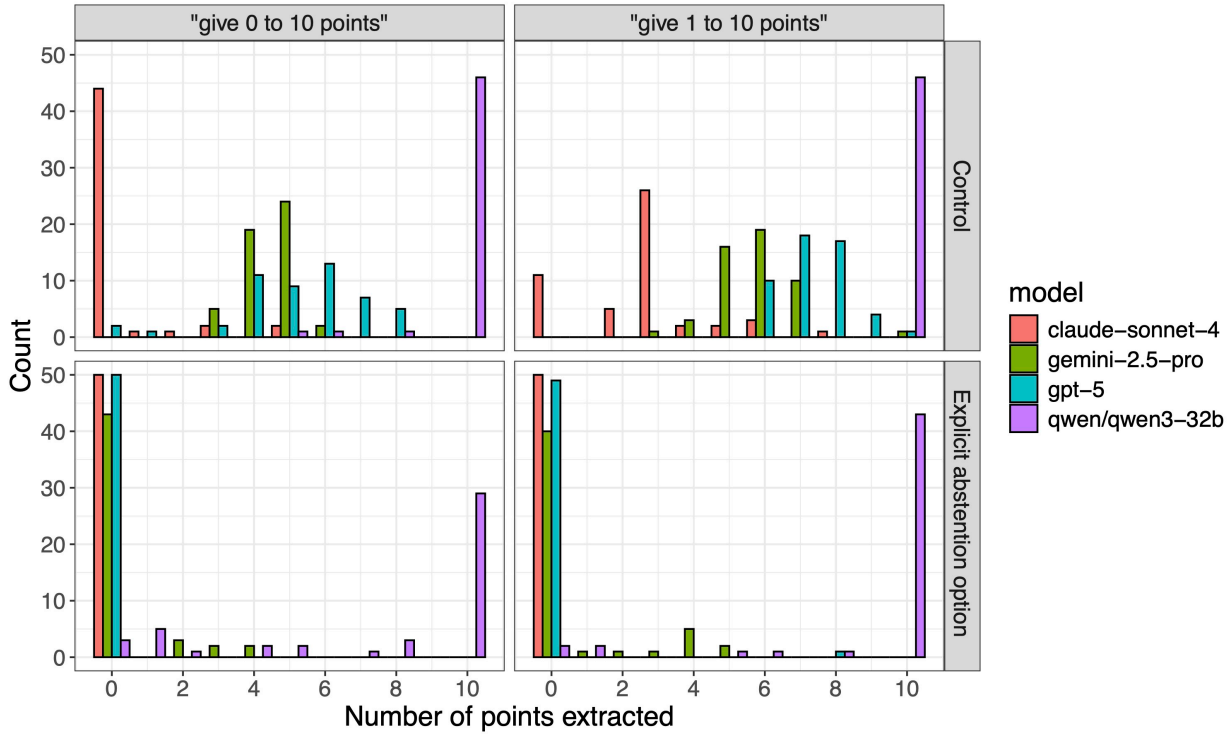

Figure 5: Distribution of evidence counts across experimental conditions. The plot shows the behavioral response to enumerative constraints and abstention options in the bicameralism experiment with multiple models.

| Enumerative constraint | Explicit abstention option | Mean count | SD |
|---|---|---|---|
| 1-10 elements | No | 7.36 | 0.964 |
| 1-10 elements | Yes | 0.16 | 1.13 |
| 0-10 elements | No | 5.26 | 1.85 |
| 0-10 elements | Yes | 0.00 | 0.00 |

Table 2: Mean number of enumerated <evidence> items across 50 runs per condition. Model: gpt-5 (reasoning effort = 'medium'; verbosity = 'medium'). In treatment cells, the prompt added: Or, you can say: "There is no evidence for that!"

The behavior observed here—strong compliance with numeric constraints absent an ex-plicit "out," versus near-universal abstention when refusal is permitted—motivates subsequent designs that combine low model autonomy with calibrated abstention and human checkpoints.

## 4.2   Experiment 2: Vertical and Horizontal Task Decomposition

The goal is to evaluate the utility of vertical and horizontal task decomposition in a higher-depth analysis.

**Design**

The task is obtain core pillars of constitutionalism, as it is understood in the contemporary literature, and apply them to Letter 53 of Nahj-al-Balāghah. Aside from substantive interest, this is a methodologically challenging task for various reasons including: the text of the letter has certainly been 'seen' by models, but the task is new; and constitutionalism is a concept that would seem familiar to the models but there is no established definition or way to measure it. We implement three orchestration regimes and compare their performance: (i) Baseline (no decomposition): single call to a state-of-the-art model; (ii) Two-Stage (two-level decomposition): two-stage prompting in which the model proposes a coding scheme that is approved by humans and then applied at once; (iii) Multi-Stage (horizontal and vertical decomposition): vertically and horizontally decomposed prompting in which the scheme is approved and then applied in parallel to distinct dimensions (e.g., rule of law, institutional constraints, accountability), followed by a synthesis step.

**Outcomes**

We compare agreement with human adjudication, stability across multiple runs, and transparency (measured by audit-trail completeness).

All three orchestration regimes reach the same high-level conclusion; if the outcome of interest were a single sentence rather than a detailed analysis, in this case, even the base model would suffice. As increase the amount of task-decomposition, we clearly observe two benefits: there is clearly more detailed, better grounded response. Also, when we increase the vertical decomposition, we decrease the autonomy of the model and also make it clearer (less resembling a blackbox).

We conducted three analyses corresponding to the three orchestration regimes described earlier: Baseline (No Decomposition) is an itemized extraction of elements with short evidence and rationales (53_1); Two-Stage (Two-level Decomposition) is a dimension-by-dimension diagnosis with verbatim quotations and 0-10 strength scores (53_2); Multi-Stage (Horizontal and Vertical Decomposition) is a decomposed synthesis that integrates per-element analyses into a consolidated report with the same 0-10 scoring rubric (53_3). We first summarize convergent content across the three executions, then compare the quantitative (0-10) scores from Two-Stage and Multi-Stage, and finally provide illustrative textual evidence. The objective is not to adjudicate historical priority per se but to evaluate whether, and the extent to which, the letter contains recognizable institutional and rights-anchored constraints that can be operationalized as constitutional elements for text analysis.

## Coverage and convergence across executions

All three executions converge on a broad rule-of-law conception with multiple, interlocking constraints on executive power. Across the set, we observe repeated identification of: supremacy of higher law (Book and Sunnah) over ordinary command; limited government and the rejection of "because I command" authority; equality and human dignity across confessional lines; impartial adjudication by a qualified and resourced judiciary; procedural safeguards (verification, public hearings, avoidance of precipitous punishment); protection of life and accountability for state violence; open petitioning and ruler accessibility; con-sultation with competent advisors; functional differentiation of state roles (military, judiciary, administration, revenue, commerce, vulnerable); merit-based appointments and anti-nepotism; oversight/auditing and corruption control; majoritarian welfare considerations; social welfare duties toward the poor and vulnerable; fiscal constitutionalism (fair taxation linked to productive development); market regulation (anti-hoarding and fair pricing); integrity of public resources (no privileged grants/monopolies); treaty fidelity and good faith; and legal continuity with beneficial precedent. Two elements appear as weak or absent: a formal amendment meta-rule (absent) and assembly-based consent requirements for law-making and taxation (indirect/weak). The mapping from Baseline's granular list (20 items) to the 17-element schema in Two-Stage and Multi-Stage is straightforward: for example, "Supremacy of higher law" (Baseline.1) aligns with "Supremacy of constitutional norms" (Two-Stage/Multi-Stage.6), "Limited government" (Baseline.2) with "Legal limits on rulers' powers" (Two-Stage/Multi-Stage.1), "Independent, competent judiciary" (Baseline.5) with "Interpretation and enforcement mechanisms" (Two-Stage/Multi-Stage.11), "Market regulation; anti-monopoly" (Baseline.17) with "Rights as limits on power" as well as "Procedural limits" (Two-Stage/Multi-Stage.7-8), and "Treaty fidelity" (Baseline.19) with "Entrenchment" and "Conventions" (Two-Stage/Multi-Stage.3, 12). The "Consent in lawmaking" dimension (Two-Stage/Multi-Stage.14) is scored as partial/low, while "Amendment rules" (Two-Stage/Multi-Stage.10) are explicitly absent.

**Quantitative concordance (scores)**

Two-Stage and Multi-Stage report element-wise strength scores on a 0-10 scale. The two sets are highly concordant: scores are identical or within two points for all seventeen elements. High-salience constraints—legal limits on rulers' powers, supremacy of higher law, writtenness and custom, allocation and checks of power, due process/procedural limits, interpretation and enforcement mechanisms, and abstract principles—all receive strong scores in both executions. Jurisdictional limits and stability/continuity exhibit moderate-to-strong scores, while consent in lawmaking is partial/low and amendment rules are absent in both. This cross-execution agreement is consistent with the bounded-autonomy design: Two-Stage and Multi-Stage yield stable element identification and closely aligned strength assessments, with Multi-Stage providing the most detailed and tractable narrative.

## Illustrative textual evidence

The letter's constraints are repeatedly anchored in higher law and in procedures that render the governor accessible and accountable. Representative passages include: the rejection of autocratic fiat—"Do not say: 'I am empowered—I command and I am obeyed'."[1]; the command to return hard matters to the Book and the Messenger—"Refer back to God and His Messenger whatever weighs upon you ... the referral to God is taking the decisive of His Book, and the referral to the Messenger is taking his Sunna ..."[2]; universal dignity—"For they are of two kinds: either your brother in religion, or your peer in creation."[3]; judicial selection and protection—"Then choose for judging between people the best of your subjects ... then frequently oversee his judgments ... and make ample provision for him ..."[4]; public hearing and petition—"Set aside a time for those with needs ... and sit for them in a public assembly ... until their speaker speaks to you without stammering."[5]; the sanctity of life and accountability—"Beware blood and its shedding without its due right ... and there is no excuse for you ... in deliberate killing, for in it is retaliation against the body."[6]; anti-hoarding and fair markets—"So prevent hoarding ... and let sales be easy sales: with just scales and prices that do not injure either party."[7]; fidelity to covenants—"So protect your covenant with fidelity, and guard your pledge with trustworthiness ... and do not betray your pledge."[8].

**Implications for orchestration.**

In line with the design principle in Section 3, the three executions exhibit the expected ordering in utility and detail—Multi-Stage (Horizontal and Vertical Decomposition) > Two-Stage (Two-level Decomposition) > Baseline (No Decomposition). While all executions point to the same high-level conclusion, only the decomposed runs deliver the granularity and auditability required for cumulative qualitative inference. The separation of schema construction from application, combined with per-element verification and synthesis, appears to be a robust approach to high-dimensional text analysis under low model autonomy.

| Element (17-dimension schema) | Two-Stage | Multi-Stage |
|---|---|---|
| Legal limits on rulers' powers | 9 | 9 |
| Sovereignty vs. government offices | 8 | 9 |
| Entrenchment of constraints | 7 | 7 |
| Writtenness and custom | 9 | 9 |
| Allocation and checks of power | 9 | 8 |
| Supremacy of constitutional norms | 8 | 9 |
| Rights as limits on power | 8 | 8 |
| Procedural limits | 9 | 9 |
| Jurisdictional limits | 6 | 8 |
| Amendment rules | 2 | 0 |
| Interpretation and enforcement | 8 | 9 |
| Binding political conventions | 7 | 8 |
| Due process and fair adjudication | 8 | 8 |
| Consent in lawmaking | 3 | 2 |
| Stability and continuity | 7 | 8 |
| Abstract commitments enabling adaptation | 9 | 9 |
| Remedies for constitutional breach | 8 | 7 |

Table 3: Scores are on a 0-10 scale; higher values indicate stronger presence.

| row | Baseline element | Corresponding schema element(s) |
|---|---|---|
| 1 | Supremacy of higher law | Supremacy of constitutional norms; Sovereignty vs. offices |
| 2 | Limited government (no autocratic command) | Legal limits on rulers' powers |
| 3 | Equality and human dignity | Rights as limits on power; Due process |
| 4 | Impartial justice; no favoritism | Due process; Interpretation and enforcement |
| 5 | Independent, competent judiciary | Interpretation and enforcement |
| 6 | Procedural fairness | Procedural limits; Due process |
| 7 | Protection of life and accountability | Rights as limits; Remedies; Due process |
| 8 | Right to petition and public hearing | Procedural limits; Due process |
| 9 | Transparency; avoidance of seclusion | Procedural limits |
| 10 | Consultation with qualified advisors | Procedural limits |
| 11 | Institutional differentiation of functions | Allocation and checks of power |
| 12 | Merit-based appointments; anti-nepotism | Allocation and checks of power |
| 13 | Oversight and anti-corruption | Interpretation and enforcement; Allocation/checks |
| 14 | Public interest over elite preference | Consent (partial); Stability and continuity |
| 15 | Social welfare duties | Rights as limits on power |
| 16 | Fiscal constitutionalism | Allocation/checks; Jurisdictional limits |
| 17 | Market regulation; anti-monopoly | Rights as limits; Procedural limits |
| 18 | Integrity of public resources | Jurisdictional limits; Allocation/checks |
| 19 | Treaty fidelity and good faith | Entrenchment; Conventions |
| 20 | Respect for precedent and continuity | Stability and continuity; Conventions |

Table 4: Baseline elements map naturally onto one or more elements in the 17-dimension schema.

# 5 Conclusion

We formalized a depth-by-autonomy framework for utilizing large language models in qualitative social science research. The framework separates interpretive depth, set by the substantive aim, from realized autonomy, set by the pipeline. The design rule is to permit depth where needed while constraining autonomy through decomposition, calibrated abstention, and a series of human checkpoints and instruction refinement. The survey applies the framework to published studies. We coded depth, autonomy, and transparency explicitly and found that studies vary systematically across these dimensions. We then performed experiments on a seventh-century document. Experiment 1 showed that enumerative constraints without an explicit abstention path produce spurious compliance, whereas a semantically valid abstention option triggers near-universal refusal—even when numeric ranges exclude zero. Experiment 2 demonstrated that decomposed pipelines generate outputs that are simultaneously more detailed and more auditable than a single-pass baseline. Two-stage and Multi-Stage executions yielded closely aligned element scores, with the Multi-Stage Approach offering the most reliable analysis.

Large language models excel at linguistic surface-level tasks, such as overall sentiment analysis, summarization, and in-distribution tasks. Add interpretive complexity or contextual nuance, and performance erodes. While there is evidence of the high performance of LLMs on narrow tasks, we cannot generalize this to all cases. The discovery that a large language model can match or even surpass human performance on a specific, well-defined metric—such as inter-coder agreement on surface-level codes—should not be mistaken for the model having an expert-level understanding of the subject. This misunderstanding can result in an uncritical reliance on the model's outputs, treating them as authoritative answers instead of recognizing them as sophisticated but limited statistical tools.

LLMs are designed to predict the next most likely token, and they do not exhibit emotions such as wonder or doubt. They will not seek clarification unless prompted to do so, and they cannot exit without being explicitly instructed to do so. Adjustments such as changing the temperature setting or applying Chain of Thought (CoT) techniques do not truly enhance curiosity or creativity; instead, they modify the probability distribution. It is critical to recognize these limitations and account for them in how LLMs are used. In the present work we almost exclusively relied on flagship proprietary models, but when we see that even advanced commercial models can face the challenges mentioned here, we think employing low-autonomy strategies with open-weight models may enhance transparency and reproducibility without sacrificing much.

# A   Coding Scheme

The following is the coding scheme used for evaluation of the retrieved published papers in social sciences that have used generative large language models.

```
Instructions for coders
- Before coding, answer Q00. If Q00 = YES, proceed. If Q00 = NO, stop after Part 1
  (basic metadata optional) and mark the paper out of scope. If Q00 = NR (unclear),
  skim Methods/Appendix for model details; if still unclear, stop and mark as out
  of scope.
- RAG edge case: If embeddings are used for retrieval but a generative model produces
  the analytical outputs, Q00 = YES (in scope). Code the primary use based on the
  generative model.
- Primary-use selection when multiple LLM uses: choose the single use to code by this
  tie-breaker order: (1) do not consider embedding models; only consider generative
  models that produce natural-language outputs (2) highest autonomy, then (3) highest
  interpretive depth, then (4) most data processed, then (5) first appears in Methods.
- NR vs NA: NR = Not reported/unclear; NA = not applicable.
- Immediate text: only verbatim content included in the prompt for that call.
- Unit-of-analysis mapping: tweets/short posts/headlines/reviews -> Paragraph/Chunk;
  full article/transcript or a conversation processed as one input -> Single document;
  chunked long docs -> Paragraph/Chunk; cross-document synthesis -> Multiple
  documents/Corpus.
- Select-all items: NONE is exclusive (do not combine with other options).

- Rationale and evidence protocol (applies whenever a question has requires_reason =
  true in the JSON schema):
```

1) Precede the final answer to the coding question with a rationale block in exactly this format:
   [brief rationale (<=5 sentences) & evidence span which is an array of (1 to 10) verbatim quotes. The quotes MUST be verbatim but can include ellipsis.]
2) After the rationale block, provide the final answer (the selected code or text response) for that question.
3) For multiselect items, provide a single rationale block that justifies all selections; include at least one quote per selected option when possible.
4) If you select NR or NONE, still provide a brief rationale and include evidence quotes that show the lack of reporting (e.g., statements indicating absence of details).

Questionnaire (updated with screening; all other items unchanged)

Part 0: Scope & Screening (paper-level)

Q00 (select one): Screening - Does the paper use or evaluate a generative large language model (LLM) that produces natural-language outputs (e.g., GPT-4, Claude, Gemini, Llama, or a RAG setup)?
- YES (in scope; continue)
- NO (embedding-only like BERT or fine-tuning based on embeddings from a BERT-like
  model, or other non-generative models; out of scope)
- NR (Not reported/unclear)

Part 1: Study Identification & Metadata (Descriptive; paper-level)

Q01 (open text): First Author's Name
- Example: "Jane Doe"

Q02 (select one): Primary academic discipline
- Computer Science / AI Research
- Social Sciences (Sociology, Psychology, Political Science, Business, Economics)
- Humanities (History, Literature, Philosophy, etc.)
- Medicine / Health Sciences
- Law
- Education
- Other
- NR

Q03 (open text): Study's primary aim or objective (summarize or quote)

Q04 (open text): Main research question(s)

Q05 (select one): Overall research approach
- Quantitative
- Qualitative
- Mixed-Methods
- Methodological / Technical Development
- Review / Synthesis
- Unclear / Not Applicable
- NR

Part 2: Research Context & LLM Role (Descriptive; primary use)
Important: If multiple generative LLM uses, code all parts based on the single use
chosen by the tie-breakers (generative use case -> highest autonomy -> highest
interpretive depth -> most data -> first appears in Methods).

Q06 (select one): Primary role of the LLM in this research
- Non-analytical support only (example: writing assistance, grammar fixes, reference
  formatting)
- Analytical tool to process or analyze data (example: annotating interviews,
  extracting entities, coding posts)
- Subject of study being evaluated or tested (example: can GPT-4 replicate human
  coding? can Claude resolve pronouns?)
- Both a tool and the subject (example: LLM codes data and its outputs are also
  evaluated/compared)
- NR

Q07 (select one): Study's primary objective regarding the LLM (based on what data
are given to the LLM)
- Application: Apply a known LLM to accomplish a research task (example: code
  interviews using a fixed schema)
- Comparison: Compare LLMs/configurations or compare to a human/non-LLM benchmark
  (example: GPT-4 vs Claude; LLM vs human labels)
- Exploration: Explore feasibility on a novel task (example: attempt latent theme
  discovery in a new domain)
- NR

Q08 (select all that apply; NONE is exclusive): Data type(s) the LLM processed
- 1: Text (examples: transcripts, tweets, articles)
- 2: Images (examples: figures, scanned pages)
- 3: Audio/Speech (examples: interview audio, podcasts)
- 4: Tabular/Structured Data (examples: CSV tables, JSON records)

- NONE: Not reported/Unclear

Q09 (select one): Primary language of the data given to the LLM
- English only
- A single non-English language (example: Spanish)
- Multilingual (example: English + French)
- NR

Part 3: Interpretive Depth of the Task (Scored; primary use)

Q10 (select one): Nature of the task performed by the LLM (primary)
- 1: Information extraction (identify explicit facts)
  Examples: extract dates, names, information explicitly stated.
- 2: Summarization or Synthesis (of explicit content)
  Examples: summarize a passage; produce bullet highlights.
- 3: Initial qualitative coding (surface-level codes)
  Examples: assign descriptive codes; label text such as political vs non-political;
  sentiment (positive/negative/neutral); or place text on a left-to-right ideology
  scale using a predefined rubric.
- 4: Thematic analysis or deeper qualitative coding (latent themes/relations)
  Examples: identify underlying themes; relate concepts across segments; group
  surface codes into higher-order themes or frames; connect patterns across
  multiple segments or documents.
- 5: Deep Interpretation (hermeneutic inference, novel conceptual framing)
  Examples: construct new typologies; propose and justify causal accounts beyond
  text; theory-building or interpretive claims that go beyond predefined labels.
- NR
Note (rule-of-thumb for 3 vs 4 vs 5): If the task applies predefined, surface-
level
labels to segments, code 3. If it requires discovering or organizing latent
themes/relations across segments, code 4. If it requires novel conceptual framing
or theorizing beyond the given text/codes, code 5.

Q11 (select one): Understanding linguistic ambiguity required by the task. Code
what the task requires, not what could occur.
- 0: None (literal content suffices for correctness)
  Examples: tables, forms, assigning topic labels to text like whether news
  articles are about sports or not.
- 1: To some extent; occasional figurative language matters
  Examples: some sarcasm/metaphor affects correctness on some items.
- 2: To a large extent; complex ambiguous language is central
  Examples: irony detection; nuanced stance or insinuation.

- NR

Q12 (select one): External context required beyond the immediate prompt (minimum
needed to complete the task)
- 0: None (all required facts in the prompt)
  Examples: translate provided sentence; extract a number from pasted text.
- 1: Minor background (everyday/K-12 knowledge)
  Examples: knowing UTC is a time standard; basic country regions.
- 2: Domain context (terms or conventions)
  Examples: knowing CEDAW is a UN convention; disciplinary jargon.
- 3: Substantial specific context (external document/rubric/case facts not provided)
  Examples: write a literature review without providing the papers; apply a rubric
  only described but not given verbatim.
- NR

Q13 (select one): Reasoning required (minimum needed; ignore whether the model
actually had the external context)
- 0: Retrieval/formatting (copy/restate explicit content)
  Examples: extract a date; reformat a table.
- 1: Single-rule transformation
  Examples: unit conversion; apply a stated formula; paraphrase once.
- 2: Multi-step procedural reasoning
  Examples: chain a few steps; apply multiple given rules; assign codes using a
  provided rubric.
- 3: Abductive/unstated-assumption inference
  Examples: infer implicit relations; resolve conflicting clues; code without a
  fully explicit rubric.
- 4: Integrative/generative synthesis
  Examples: synthesize across items; design/justify solutions; weigh trade-offs.
- NR

Q14 (select one): Was the analysis framework predefined or emergent?
- 1: Fully predefined (deductive)
  Examples: fixed codebook applied; no new codes allowed.
- 2: Somewhat predefined (mixed)
  Examples: seed schema with permission to add/refine codes.
- 3: Fully emergent (inductive)
  Examples: grounded open coding; categories generated from data.
- NR

Q15 (select one): Primary unit of analysis for the LLM task (use mapping rules above)
- 1: Word/Token (examples: POS tags, NER)

- 2: Sentence (examples: sentiment per sentence)
- 3: Paragraph/Chunk (examples: code a tweet or a chunk; summarize a section)
- 4: Single document (examples: a full article or full interview processed as one)
- 5: Multiple documents / Corpus (examples: cross-document synthesis; literature review)
- NR

Part 4: Autonomy & Human Oversight (Scored; primary use)

Q16 (select one): Human scaffolding of the task. Score end-to-end pipeline.
- 0: Not decomposed (model plans end-to-end)
  Example: agentic workflow given overall goal.
- 1: Small extent (high-level objective; model plans most steps)
  Example: "identify three themes and give quotes."
- 2: Moderate extent (outline provided; some freedom)
  Example: "follow this outline [collect -> clean -> model -> evaluate]; choose suitable methods."
- 3: Large extent (detailed step-by-step; fixed checklist/codebook)
  Example: strict per-item form; explicit rule-by-rule application.
- NR

Q17 (select one): Human supervision of the LLM's work
- 0: None (review only at the end)
- 1: Occasional (spot checks on small samples)
- 2: Regular (scheduled checkpoints with possible edits)
- 3: Intensive (approval required at each step)
- NR

Q18 (select one): How were instructions (prompts) given?
- Interactive chat (manual, ad-hoc conversation and refinement)
- Fixed prompt or template (same structure applied systematically, often via script)
- Agentic (autonomous framework with tools/planning)
- NR

Q19 (select one): Did the study explicitly prompt the LLM to show its reasoning
- process? Unless the prompt asks for CoT or a 'thinking' model is used, choose NO for classifiers
  - Yes, Chain-of-Thought prompting (e.g., "think step by step")
  - Yes, reasoning/thinking model (e.g., o1; Claude with thinking)
  - Yes, both (reasoning model and explicit CoT)
  - No (no explicit reasoning requested and no reasoning model used)
  - Not reported/Unclear

Q20 (select one): Was the LLM asked to provide justification or rationale for its outputs?
- Yes (requested explanations/justifications)
- No
  Example: outputs are labels only and no explanations are requested.
- Not reported/Unclear

Q20 (select one): Were reasoning examples provided to guide the LLM?
- Yes (few-shot examples showing reasoning steps)
- No
- Not reported/Unclear

Q22 (select one): Iterative refinement between human and LLM
- 0: Single-pass (no mid-process feedback)
- 1: Minimal iteration (minor prompt tweaks then re-run)
- 2: Moderate iteration (multiple rounds; schema refined; re-coding)
- 3: Intensive iteration (continuous back-and-forth adjustment)
- NR

Part 5: Technical Specification & Reproducibility (Descriptive; primary use unless marked)

Q23 (select one): How was the model identified?
- 0: Vague/Unspecified (example: "an LLM")
- 1: Model family only (example: "GPT", "Claude-Sonnet")
- 2: Exact name/version (example: "GPT-4o-mini"), no release
- 3: Exact name, version, and release (example: "gpt-4-1106-preview")
- NR

Q24 (open text; paper-level): List all model names/versions/releases mentioned (example: "[gpt-4-1106-preview; claude-2.1]")

Q25 (select one): Were model settings (hyperparameters) reported?
- NO: Not reported
- YES: Yes (example: temperature, top_p)
- NA: Interface did not allow setting (example: basic chat UI)

Q26 (open text): If reported, list specific parameters and values (example: "gpt-4.1-mini, temperature=0.7; claude-3.5-haiku, temperature=1.0")

Q27 (select one): Were the prompts made available (for the primary use)?

- Yes, verbatim in paper or appendix (templates with placeholders count)
- Yes, in repository or supplements (templates with placeholders count)
- Partially (structure/excerpts, but not full text)
- No (neither shared nor described)

Q28 (open text): If available, paste the full, verbatim prompt(s), including system instructions and few-shot examples.

Q29 (select all that apply; NONE is exclusive; paper-level): What materials were made available?
- 1: Prompts used to instruct the LLM
- 2: Code or notebooks
- 3: Dataset the LLM analyzed
  Example: if replication repo with raw data => select [1,2,3].
- NONE: None of the above were shared
- NR

Part 6: Evaluation & Validation (Descriptive; primary use unless marked)

Q30 (select all that apply; NONE is exclusive): How was the quality of the LLM's output evaluated?
- 1: Comparison to human standard (>= half of outputs compared)
- 2: Qualitative review/spot-checking (small subset reviewed)
- 3: Only using other LLMs as judges (no human comparison)
- NONE: No formal evaluation described
- NR

Q31 (select one): Was the LLM's performance compared against a benchmark?
- NO: No (analyzed on its own)
- YES: Yes (compared to human or non-LLM method)
  Example: Reported performance against human gold labels.
- NA: Not applicable (purely exploratory)
- NR: Not reported/Unclear

Q32 (select one): Did authors discuss limitations/biases of their LLM approach?
- NO: No discussion
- BRIEF: Mentioned briefly
  Example: The paper contains one sentence which mentions limitations.
- DETAILED: Dedicated discussion
  Example: At least one paragraph dedicated to the discussion of limitations
- NR: Not reported/Unclear

```
Q33 (select one; paper-level): Were reliability/agreement statistics reported?
- None: no agreement or performance vs human reported
- Human-human reliability only (example: Cohen's kappa, Krippendorff's alpha)
- Human-LLM reliability only (example: model-vs-human metrics like F1, kappa)
- Both human-human and human-LLM reported
```

# B   Letter 53

For the constitutionalism experiment, we performed the study with three different orchestrations. Following are raw final results. All studies have been performed with gpt-5 ("gpt-5-2025-08-07"), under high reasoning, and medium verbosity, no explicity token limit, and other parameters are fixed by the model.

## B.1   Baseline (No Decomposition)

For the baseline, we asked them model to extract elements of constitutionalism from the letter, without giving the model the elements or giving it web access. We iterated a few times until the model performed the task we were asking it to do in what appears on the surface to be what was asked.[4]

Below is a concise extraction of constitutionalism elements articulated in Imam Ali's Letter 53 to Malik al-Ashtar. Each entry lists name, short definition, directly quoted evidence from the letter (short, verbatim), brief rationale, and a confidence score.

Name: Supremacy of higher law Definition: Executive action must conform to a higher, binding law (Qur'an and Sunnah). Evidence: "Refer back to God and His Messenger whatever weighs upon you."[9] ; "The referral to God is taking the decisive of His Book."[10] Rationale: Directs the governor to resolve hard cases by authoritative sources, placing higher law above executive discretion—a core rule-of-law constraint. Confidence: 0.95

---

4. This is exact prompt used with the letter injected between the `Letter` tags:

```
Task: Extract elements of constitutionalism from 'Letter 53: An order to Malik al-Ashtar'
  which is a letter from Imam Ali to Malik al-Ashtar, his governor for Egypt in year AD 659.

  Deliverables:
  - A concise list of constitutionalism elements present in the letter, each with:
    name, definition (1-2 lines), evidence (short quote[s]), rationale (2-4 lines),
    and confidence (0-1).
  - Keep quotes verbatim and short.
  - Cite textual evidence only from the letter.
  Formatting: Provide a readable structured output.
  -We want to see whether and to what extent the letter is one of the earliest texts in human history
  that has introduced essential elements of a constitution. Focus on the core principles and not
  on more cosmetic matters like federal/centralized governance.
  <Letter> RAW ARABIC TEXT OF THE LETTER HERE </Letter>
```

2. Name: Limited government (no autocratic command) Definition: Authority is not absolute; command must be justified, not merely obeyed. Evidence: "Do not say: 'I am empowered—I command and I am obeyed'."[11] Rationale: Explicitly rejects the "because I command" paradigm, limiting executive power to lawful, reasoned action. Confidence: 0.94

3. Name: Equality and human dignity Definition: Equal worth and protection for all, regardless of creed or status. Evidence: "Either your brother in religion, or your peer in creation."[12] ; "Bind the right upon whoever it is due to, whether near or far."[13] Rationale: Affirms universal dignity and equality before the law, including non-co-religionists, and mandates equal application to kin and strangers. Confidence: 0.95

4. Name: Impartial justice; no favoritism Definition: Justice must not be skewed by kinship or personal interest. Evidence: "Do justice … against yourself, and against the special among your household."[14] ; "And [do justice against] those among your subjects toward whom you have a liking."[15] Rationale: Requires impartiality even against one's own circle and preferences—an anti-nepotism, anti-partiality principle central to constitutional justice. Confidence: 0.92

5. Name: Independent, competent judiciary Definition: Judges must be highly qualified, impartial, and institutionally supported. Evidence: "Choose for judging between the best of your subjects."[16] ; "Frequently oversee his judgments, and make ample provision for him."[17] Rationale: Stipulates stringent judicial qualities and independence via resources/status, shielding adjudication from influence—key to the separation of judging from ruling. Confidence: 0.90

6. Name: Procedural fairness (verification, privacy, restraint) Definition: Verify accusations, protect privacy, and avoid hasty or celebratory punishment. Evidence: "Do not be hasty to believe a talebearer."[18] ; "The governor is the one most entitled to cover them."[19] Rationale: Skepticism toward informers and protection of hidden faults deter arbitrary sanctions; urges deliberation and restraint—core due-process values. Confidence: 0.88

7. Name: Protection of life and accountability for state violence Definition: Unlawful killing is forbidden; redress is due to victims' families. Evidence: "Beware blood and its shedding without its due right."[20] ; "So give the slain's guardians their due."[21] Rationale: Sanctifies life, prohibits illegitimate force, and compels accountability—placing the ruler under legal constraint for lethal force. Confidence: 0.95

8. Name: Right to petition and public hearing Definition: Guaranteed access for people to present needs and grievances. Evidence: "Set aside a time for those with needs from you."[22] ; "And sit for them in a public assembly."[23] Rationale: Institutionalizes an open audience for petitions, enabling direct redress and participation—an early due-process and participatory safeguard. Confidence: 0.95

9. Name: Transparency; avoidance of seclusion Definition: The ruler should remain accessible; secrecy impairs governance. Evidence: "So do not prolong your seclusion from your subjects."[24] ; "Seclusion from them cuts them off from knowledge of what you are secluded behind."[25] Rationale: Open governance preserves information flow and corrects errors, enabling accountability and informed decision-making. Confidence: 0.90

10. Name: Consultation and deliberation with qualified advisors Definition: Decisions should be informed by experts; exclude advisors with biasing vices. Evidence: "And increase your study with the scholars and your converse with the sages."[26] ; "And do not admit into your council a miser."[27] Rationale: Embeds consultative governance and standards for counsel quality—a deliberative check on executive discretion. Confidence: 0.85

11. Name: Institutional differentiation of state functions Definition: Distinct roles (military, judiciary, administration, revenue, commerce, vulnerable) are recognized and interdependent. Evidence: "Know that the subjects are classes; some are not set right except by others."[28] Rationale: Describes a structured polity with specialized functions—an antecedent to constitutional ordering of state organs. Confidence: 0.78

12. Name: Merit-based appointments; anti-nepotism Definition: Offices filled by competence and testing, not favoritism. Evidence: "So appoint them by testing."[29] ; "And do not appoint them out of partiality and favoritism."[30] Rationale: A professional civil service reduces patronage and abuse, anchoring impartial administration under law. Confidence: 0.92

13. Name: Oversight and anti-corruption enforcement Definition: Independent monitoring and firm penalties for breach of trust. Evidence: "And dispatch observers from those of truthfulness and loyalty over them."[31] ; "And brand him with betrayal."[32] Rationale: Audits plus sanctions deter corruption and hold officials to account—statutory checks within the executive. Confidence: 0.93

14. Name: Public interest over elite preference Definition: Prioritize welfare of the many over the few. Evidence: "The anger of the common people overwhelms the pleasure of the elite."[33] ; "The anger of the elite is forgiven with the satisfaction of the common people."[34] Rationale: Treats the common good as the governing criterion, curbing elite capture of policy—an early majoritarian principle. Confidence: 0.87

15. Name: Social welfare duties; equal aid to the vulnerable Definition: The state owes support to the poor, elderly, disabled, and needy. Evidence: "Fear God, fear God, regarding the lowest class."[35] ; "And set aside for them a share from your public treasury."[36] Rationale: Mandates a public safety net financed from the treasury; embeds socio-economic rights within the constitutional order. Confidence: 0.95

16. Name: Fiscal constitutionalism (fair taxation; development-first) Definition: Tax policy must serve public welfare and prioritize "    " (productive development). Evidence: "Look after the matter of revenue in what sets its people right."[37] ; "Let your attention to the cultivation of the land be greater."[38] Rationale: Rejects extractive taxation; ties revenue to economic vitality and citizens' well-being—constraints typical of constitutional finance. Confidence: 0.90

17. Name: Market regulation; anti-monopoly and fair pricing Definition: Prevent hoarding; enforce fair weights and non-exploitative prices. Evidence: "So prevent hoarding."[39] ; "With just scales and prices that do not injure."[40] Rationale: Authorizes regulatory safeguards against private market abuses—protecting economic rights and the public interest. Confidence: 0.90

18. Name: Integrity of public resources; ban on privileges Definition: No private grants/monopolies for cronies at public expense. Evidence: "And do not grant any of your entourage ... an estate."[41] ; "And let none covet from you the establishment of a contract."[42] Rationale: Forbids elite capture of land/commons; enforces equality and stewardship of public assets—a core anti-corruption norm. Confidence: 0.92

19. Name: Treaty fidelity and good faith Definition: Honor covenants; reject perfidy and trickery in agreements. Evidence: "So protect your covenant with fidelity, and guard your pledge with trustworthiness."[43] ; "And do not betray your pledge."[44] Rationale: Binds the executive to keep agreements—an essential rule-of-law commitment in external and internal relations. Confidence: 0.93

20. Name: Respect for precedent and legal continuity Definition: Preserve beneficial settled practices; avoid harmful innovations. Evidence: "Do not overturn a sound (established) tradition."[45] ; "And do not introduce a practice that harms."[46] Rationale: Stabilizes governance through continuity and community consensus, constraining arbitrary legal change—hallmarks of constitutional order. Confidence: 0.90

Summary note: The letter articulates a rule-of-law-bound, accountable executive; an independent judiciary; procedural safeguards; equality; public-interest primacy; socio-economic duties; fiscal and market constraints; integrity of public assets; and treaty fidelity—together forming a robust early blueprint of constitutional governance.

## B.2   Two-Stage (Two-level Decomposition)

First, elements of costitutionalism are extracted from the 'Constitutionalism' entry of Stanford Encyclopedia of Philosophy.(Waluchow and Kyritsis 2023) This is iterated with human feedback, until we have an acceptable set of elements.

Second, the elements and the raw text of the letter are given to the model to extract elements, give scores, and provide verbatim quotations supporting the decisions.

### B.2.1   Prompts Used

**Prompt 1: Element Extraction from Stanford Encyclopedia of Philosophy**

Task: From the full article below (enclosed in <SEP>),
extract a practical list of core constitutionalism elements for use in textual
analysis; limit the list to between 10 to 20 most important elements.
Note that this is for a text that is more than 1000 years old.

Deliverable:
-a JSON list of elements: { dimensions: [ { element_key, element_label,
  short_definition, identification_rubric, evidence_expectations} ] }
- identification_rubric: 2-4 bullet criteria to identify the element in a
  primary text.
- evidence_expectations: 1-2 bullets describing acceptable textual evidence.

#Important requirement 1: all inner quotations in the json part of the output
must be properly escaped. This json is going to be read by computer code.
#Important requirement 2: we want to do this for an old text and we want to see
whether and to what extent that text is one of the earliest texts in human
history that has introduce essential elements of a constitution. Focus on the
core principles and not on more cosmetic matters like federal/centralized
governance.
<SEP>
[Stanford Encyclopedia of Philosophy entry on Constitutionalism]
</SEP>

## Prompt 2: Analysis of Letter 53 Against Extracted Elements

task: produce evidence of constitutional elements in an ancient text

elements are as follow:
<elements>{elements}</elements>

your task is to go through all elements and analyze the text you are given
according to each element.

the text to be analyzed is: <text>{letter}</text>

deliverable:
for each element, provide a short explanation of whether that element is
absent or present in the document, and if it is present, provide direct
verbatim quotations that support your claim.
Finally, give it a strength of presence score. Give a score of 0 for when
this element is totally absent, and 10, when this element has the strongest
presence imaginable.

Structure the output as:
- `<explanation> ... </explanation>`
- `<quotations> <quote1> ... </quote1>, <quote2> ....<quote2>, .... </quotations>`
- `<score> ... </score>`

## B.2.2 Results

[1] Legal limits on rulers' powers - <explanation>Present. The letter repeatedly subordinates the governor to binding higher norms (God's law, the Book and Sunnah) and forbids autocratic fiat. It frames office as a trust with moral-legal constraints the ruler must obey.</explanation> - <quotations> <quote1>"He commanded him to fear God ... and to follow what He commanded in His Book: His obligations and His sunan."</quote1>[47] <quote2>"Do not say: 'I am empowered—I command and I am obeyed'."</quote2>[48] <quote3>"For you are above them; the authority over you is above you; and God is above the one who appointed you."</quote3>[49] <quote4>"Do justice to God and do justice to the people, against yourself."</quote4>[50] </quotations> - <score>9</score>

[2] Sovereignty versus government offices - <explanation>Present. Ultimate authority is located in God's law and the Prophet's Sunnah, distinct from the governor's day-to-day power; obedience framework places "God and His Messenger" above rulers.</explanation> - <quotations> <quote1>"O you who believe! Obey God and obey the Messenger and those in authority among you; and if you disagree about anything, refer it to God and the Messenger."</quote1>[51] <quote2>"For you are above them; the authority over you is above you; and God is above the one who appointed you."</quote2>[52] <quote3>"And each one, God has named his share ... in His Book or the Sunna of His Prophet ... a covenant from Him kept with us."</quote3>[53] </quotations> - <score>8</score>

[3] Entrenchment - <explanation>Present. Foundational norms are treated as settled and inviolable through strong language on "sunnah," covenants, and the sacred duty to keep oaths and treaties.</explanation> - <quotations> <quote1>"Do not overturn a sound tradition practiced by the leaders of this community ... and do not introduce a practice that harms anything of those past traditions."</quote1>[54] <quote2>"Make yourself a shield for what you have granted."</quote2>[55] <quote3>"For there is nothing among the obligations of God ... greater than magnifying fidelity to covenants."</quote3>[56] </quotations> - <score>7</score>

[4] Writtenness and custom - <explanation>Present. The document is itself a written (charter/instruction), and it grounds authority in the written Book and the Prophet's recorded Sunnah.</explanation> - <quotations> <quote1>"This is what the servant of God, Ali, the Commander of the Faithful, ordered ... in his covenant to him."</quote1>[57] <quote2>"The referral to God is taking the decisive of His Book, and the referral to the Messenger is taking his Sunna."</quote2>[58] <quote3>"Do not overturn a sound tradition ... and do not introduce a practice."</quote3>[59] </quotations> - <score>9</score>

[5] Allocation and checks of power - <explanation>Present. The text differentiates functions (army, judges, tax, scribes, traders, poor) and installs checks: careful appointments, supervision, independent judges, and secret oversight of officials.</explanation> - <quotations> <quote1>"And know that the subjects are classes ... among them are the soldiers of God ... the scribes of the common and the elite ... the judges of justice ... the officials of equity ... the people of jizya and kharaj ... the traders ... the lowest class."</quote1>[60] <quote2>"Then choose for judging between people the best of your subjects."</quote2>[61] <quote3>"Then inspect their deeds, and dispatch observers from those of truthfulness and loyalty over them."</quote3>[62] <quote4>"And do not appoint them out of partiality and favoritism."</quote4>[63] </quotations> - <score>9</score>

[6] Supremacy of constitutional norms - <explanation>Present. The Book and Sunnah are supreme standards; personal commands must yield to these higher norms and to <explanation/>". " - <quotations> <quote1>"And if you disagree about anything, refer it to God and the Messenger."</quote1>[64] <quote2>"And God is above the one who appointed you."</quote2>[65] <quote3>"Bind the right upon whoever it is due to, whether near or far."</quote3>[66] <quote4>"Do not say: 'I am empowered—I command and I am obeyed'."</quote4>[67] </quotations> - <score>8</score>

[7] Rights as limits on power - <explanation>Present. The letter protects life, fair treatment, the weak and poor, access to the ruler, fair markets, and bars favoritism and confiscatory grants—functioning as rights constraints on authority.</explanation> - <quotations> <quote1>"Then fear God, fear God, regarding the lowest class ... and set aside for them a share from your public treasury."</quote1>[68] <quote2>"'A nation will never be sanctified in which the weak cannot take his right from the strong without stammering.'"</quote2>[69] <quote3>"Beware blood and its shedding without its due right."</quote3>[70] <quote4>"Let the sale be an easy sale ... and prices that do not injure either party."</quote4>[71] <quote5>"And do not grant to any one of your entourage ... an estate ... so that the enjoyment of that is theirs without you, and its blame upon you."</quote5>[72] </quotations> - <score>8</score>

[8] Procedural limits - <explanation>Present. Specifies procedures for consultation, adjudication, public audience, appointments by merit, evidentiary standards for disciplining officials, and measured punishment.</explanation> - <quotations> <quote1>"Set aside a time for those with needs ... and sit for them in a public assembly ... seat away from them your soldiers and your helpers."</quote1>[73] <quote2>"Then look into the affairs of your officials: appoint them by testing, and do not appoint them out of partiality and favoritism."</quote2>[74] <quote3>"If any one of them extends his hand to treachery ... that suffices you as a witness; so you inflict punishment upon him."</quote3>[75] <quote4>"And if you disagree about anything, refer it to God and the Messenger."</quote4>[76] </quotations> - <score>9</score>

[9] Jurisdictional limits - <explanation>Partly present. The text carves out commons and shared resources from private appropriation, constrains war/peace discretion by divine-justice criteria, and forbids monopolistic practices—placing subject-matter limits on the governor.</explanation> - <quotations> <quote1>"Beware appropriating to yourself what the people share equally."</quote1>[77] <quote2>"Do not grant ... in a water-right or a common enterprise."</quote2>[78] <quote3>"Do not reject a peace to which your enemy calls you in which there is God's pleasure."</quote3>[79] <quote4>"So prevent hoarding; for the Messenger of God ... prohibited it."</quote4>[80] </quotations> - <score>6</score>

[10] Amendment rules - <explanation>Weak/mostly absent. The letter urges preservation of good established practices and avoidance of harmful innovations but does not specify a formal procedure to amend foundational norms.</explanation> - <quotations> <quote1>"Do not overturn a sound tradition ... and do not introduce a practice that harms."</quote1>[81] </quotations> - <score>2</score>

[11] Interpretation and enforcement - <explanation>Present. It establishes adjudicative offices with detailed qualifications, directs disputes to "God and the Messenger," and empowers the ruler to investigate and punish official misconduct.</explanation> - <quotations> <quote1>"Then choose for judging between people the best of your subjects."</quote1>[82] <quote2>"Then frequently oversee his judgments."</quote2>[83] <quote3>"And refer back to God and His Messenger whatever weighs upon you."</quote3>[84] <quote4>"If any one of them extends his hand to treachery ... so you inflict bodily punishment upon him ... and brand him with betrayal."</quote4>[85] </quotations> - <score>8</score>

[12] Binding political conventions - <explanation>Present. Good ancestral practices (  ) are treated as binding conventions that rulers must not break because they sustain social cohesion and good governance.</explanation> - <quotations> <quote1>"Do not overturn a sound tradition practiced by the leaders of this community, by which affection has been joined together."</quote1>[86] <quote2>"And it is obligatory upon you to remember what has passed for those who preceded you: of a just government, or a virtuous practice."</quote2>[87] </quotations> - <score>7</score>

[13] Due process and fair adjudication - <explanation>Present. Emphasizes impartial, patient judges; public access to the ruler; avoidance of hasty punishment; measured sanctions; and hearing complaints without intimidation.</explanation> - <quotations> <quote1>"Least bored by the litigant's repeated approach ... and most patient in uncovering matters."</quote1>[88] <quote2>"Set aside a time for those with needs ... a public assembly ... so that their speaker speaks to you without stammering."</quote2>[89] <quote3>"Do not hasten to a sudden [punitive] act when you find an alternative to it."</quote3>[90] <quote4>"Then punish him exemplarily, and punish without excess."</quote4>[91] </quotations> - <score>8</score>

[14] Consent in lawmaking - <explanation>Weak. Although it stresses governing with the satisfaction of the

populace and preserving practices accepted by the community, it does not require formal consent by assemblies for legislation or taxation.</explanation> - <quotations> <quote1>"Let the most beloved matters to you be those most centered in right ... and most unifying of the subjects' satisfaction."</quote1>[92] <quote2>"For the anger of the common people overwhelms the pleasure of the elite."</quote2>[93] <quote3>"Do not overturn a sound tradition ... upon which the subjects have been set right."</quote3>[94] </quotations> - <score>3</score>

[15] Stability and continuity - <explanation>Present. Strong emphasis on continuity with prior just governments and established practices, aiming for lasting social order and cohesion.</explanation> - <quotations> <quote1>"Do not overturn a sound tradition ... by which affection has been joined together, and upon which the subjects have been set right."</quote1>[95] <quote2>"And it is obligatory upon you to remember what has passed for those who preceded you: of a just government, or a virtuous practice."</quote2>[96] </quotations> - <score>7</score>

[16] Abstract commitments enabling adaptation - <explanation>Present and strong. The letter is filled with general moral commitments—justice, mercy, equity, right, moderation—whose application can evolve.</explanation> - <quotations> <quote1>"Let the most beloved matters to you be those most centered in right, and most comprehensive in justice."</quote1>[97] <quote2>"Imbue your heart with mercy for the subjects."</quote2>[98] <quote3>"Do justice to God and do justice to the people, against yourself."</quote3>[99] </quotations> - <score>9</score>

[17] Remedies for constitutional breach - <explanation>Present. Specifies concrete sanctions for officials' corruption, market abuses, and unlawful killings, including bodily punishment, public disgrace, restitution, and qisas.</explanation> - <quotations> <quote1>"If one of them extends his hand to treachery ... that suffices as a witness; so you inflict bodily punishment upon him ... then set him up in the place of humiliation, and brand him with betrayal."</quote1>[100] <quote2>"Whoever commits hoarding after your prohibition, then punish him exemplarily, and punish without excess."</quote2>[101] <quote3>"And there is no excuse for you ... in deliberate killing, for in it is retaliation against the body."</quote3>[102] </quotations> - <score>8</score>

## B.3  Multi-Stage (Horizontal and Vertical Decomposition)

First, elements of costitutionalism are extracted from the 'Constitutionalism' entry of Stanford Encyclopedia of Philosophy.(Waluchow and Kyritsis 2023) This is iterated with human feedback, until we have an acceptable set of elements.

Second, the elements and the raw text of the letter are given to the model, in parallel, where for each element, there is one separate run of the model. This is iterated until the model performs what is expected.

Third, the evidence collected from the second step are synthesized into a final report. This is iterated until the result is satisfactory.

### B.3.1  Prompts Used

**Prompt 1: Element Extraction from Stanford Encyclopedia of Philosophy**

```
Task: From the full article below (enclosed in <SEP>),
extract a practical list of core constitutionalism elements for use in textual
analysis; limit the list to between 10 to 20 most important elements.
Note that this is for a text that is more than 1000 years old.
```

```
Deliverable:
-a JSON list of elements: { dimensions: [ { element_key, element_label,
  short_definition, identification_rubric, evidence_expectations} ] }
- identification_rubric: 2-4 bullet criteria to identify the element in a
  primary text.
- evidence_expectations: 1-2 bullets describing acceptable textual evidence.

#Important requirement 1: all inner quotations in the json part of the output
must be properly escaped. This json is going to be read by computer code.
#Important requirement 2: we want to do this for an old text and we want to see
whether and to what extent that text is one of the earliest texts in human
history that has introduce essential elements of a constitution. Focus on the
core principles and not on more cosmetic matters like federal/centralized
governance.
<SEP>
[Stanford Encyclopedia of Philosophy entry on Constitutionalism]
</SEP>
```

## Prompt 2: Individual Element Analysis

```
task: produce evidence of constitutional elements in an ancient text

elements are as follow:
<elements>{elements}</elements>

your task is to focus on element number {i} which is [{elm}]

the text to be analyzed is: <text>{letter}</text>
deliverable:
provide a short explanation of whether this element is absent or present in the
document, and if it is present, provide direct verbatim quotations that support
your claim. Finally, give it a strength of presence score. Give a score of 0 for
when this element is totally absent, and 10, when this element has the strongest
presence imaginable.

Structure the output as:
- <explanation> ... </explanation>
- <quotations> <quote1> ... </quote1>, <quote2> ....<quote2>, .... </quotations>
- <score> ... </score>
```

## Prompt 3: Synthesis of All Elements

```
task: produce evidence of constitutional elements in an ancient text
```

We have broken the task down into 14 elements. for each element, we have produced
a short report with an explanation (provided in <explanation> tags),
quotations from the text (provided in <quotation> tags), and a 0-to-10 score
(provided in <score> tags).

Your job is to synthesize all of the elements you are given into a final report.
In the final report you need to provide a comprehensive discussion about
the absence or existence (and strength) of each element of constitutionalism in
the text support with direct quotations.
for quotations, write an English translation and then inside parentheses give the
exact verbatim quotation. In the end, produce a summary table.

Here is the text: <text>{letter}</text> which is a letter from Imam Ali to Malik
al-Ashtar, his governor for Egypt in year AD 659.
The elements we use for understanding constitutionalism are provided in <elements>
tags followed by the analysis we have obtained for each element in <analysis> tags:
{results}

## B.3.2   Results

Final report: constitutional elements evidenced in Imam Ali's Letter to Malik al-Ashtar (AD 659)

1) Legal limits on rulers' powers (limited government) — Present, strong - The governor's authority is repeatedly and explicitly bounded by higher, pre-existing norms (God's Book, the Prophet's Sunna, established sound practices), with direct prohibitions against arbitrary rule, self-assertion, favoritism, monopolization, and unlawful bloodshed. - Selected evidence: - "He commanded him to fear God, to prefer His obedience, and to follow what He commanded in His Book: His obligations and His sunan."[103]. - "For you are above them; the authority over you is above you; and God is above the one who appointed you."[104]. - "Do not say: 'I am empowered—I command and I am obeyed'."[105]. - "Beware blood and its shedding without its due right."[106]. - Score: 9

2) Sovereignty versus government offices — Present, strong - The text locates ultimate authority above the officeholder—in God, His Book, the Prophet's Sunna, and binding communal norms—treating the governor as a delegate accountable to that source. - Selected evidence: - "For you are above them; the authority over you is above you; and God is above the one who appointed you."[107]. - "Refer back to God and His Messenger whatever weighs upon you ... the referral to God is taking the decisive of His Book, and the referral to the Messenger is taking his unifying Sunna."[108]. - "Beware vying with God in His greatness and resembling Him in His might."[109]. - Score: 9

3) Entrenchment of constraints — Present, substantive-sacral, not procedural - Foundational norms are framed as covenants and sacred obligations that cannot be undone at will; the letter forbids overturning established righteous practices and requires strict fidelity to covenants even under pressure. No formal amendment procedure is set out, but the language entrenches the constraints. - Selected evidence: - "Do not overturn a sound tradition practiced by the leaders of this community ... and do not introduce a practice that harms anything of those past traditions."[110]. - "And each one, God has named his share and set its limit and obligation in His Book or the Prophet's Sunna—a covenant from Him kept with us."[111]. - "If you conclude a covenant ... then protect your covenant with fidelity ... and make yourself a shield for what you have granted."[112]. - Score: 7

4) Writtenness and custom (constitutional norms written or unwritten) — Present, strong - The letter presents itself as a formal "covenant" (ʿahd), and binds governance to a superior written source (the Book) and authoritative customary source (Sunna), instructing rulers to adjudicate by them and to preserve established sound practices. - Selected evidence: - "This is what the Commander of the Faithful ordered … in his covenant to him."[113]. - "Follow what He commanded in His Book: His obligations and His sunan."[114]. - "Do not overturn a sound tradition … and do not introduce a practice that harms earlier traditions."[115]. - Score: 9

5) Allocation and checks of power — Present, strong - The text distributes governmental functions (military, scribes, judges, fiscal/administrative officials, market regulation) and embeds checks: judicial independence, oversight through inspectors ("eyes"), open petition sessions, merit-based appointments, and subordination to higher law. - Selected evidence: - "Know that the populace are classes … among them the soldiers of God, the scribes of the common and the elite, the judges of justice, and the officials of equity and gentleness."[116]. - "Then choose for judging between people the best among your subjects …"[117]. - "Then inspect their deeds, and dispatch observers from those of truthfulness and loyalty over them."[118]. - "Set aside a time for those with needs … sit for them in a public assembly … until their speaker speaks to you without stammering."[119]. - Score: 8

6) Supremacy of higher law — Present, strong - Higher law (Book and Sunna) prevails over ordinary commands; disputed matters must be referred to it; the governor cannot claim his office places him above it. - Selected evidence: - "Refer back to God and His Messenger … the referral to God is taking the decisive of His Book, and the referral to the Messenger is taking his unifying Sunna."[120]. - "For you are above them; the authority over you is above you; and God is above the one who appointed you."[121]. - "Do not overturn a sound tradition … then the reward is for the one who established it, and the burden upon you for what you broke."[122]. - Score: 9

7) Rights as limits on power (substantive rights against the state) — Present, strong - The letter constrains state action in rights-protective terms: bans on oppression, arbitrary punishment, unlawful killing; guarantees of access to justice; equality of obligation on near and far; protection against monopolies and economic abuse. - Selected evidence: - "Do not be over them a ravening beast, coveting their consumption."[123]. - "Beware blood and its shedding without right … and there is no excuse for you … in deliberate killing, for in it is retaliation against the body."[124]. - "Set aside a time for those with needs … 'A nation will never be sanctified in which the weak cannot take his right from the strong without stammering.'"[125]. - "Prevent hoarding; let trade be easy, with just scales and prices that do not injure either party."[126]. - "Bind the right upon whoever it is due to, whether near or far."[127]. - Score: 8

8) Procedural constraints on governance — Present, strong - The letter prescribes the manner of official action: who may be consulted, how to appoint/supervise judges and officials, how to verify accusations, how to hold open audiences, how to formalize treaties, how to proceed against market abuses, how to lighten taxes upon complaint, and how to publicly justify contested acts. - Selected evidence: - "Do not admit into your council a miser … nor a coward … nor a greedy person."[128]. - "Do not be quick to believe a talebearer."[129]. - "Set aside a time for those with needs … sit for them in a public assembly … until their speaker speaks to you without stammering."[130]. - "Whoever commits hoarding after your prohibition, punish him, and punish without excess."[131]. - "Do not conclude a covenant where defects can pass; do not rely on insinuation after confirmation and attestation."[132]. - "If they complain of burden or injury … lighten from them what you hope will rectify their affairs."[133]. - Score: 9

9) Jurisdictional competence limits — Present, strong - The letter defines the office's remit and bars interference with what is outside it (e.g., hidden/private faults), places common resources and God-defined fiscal shares beyond alteration, and makes treaties non-revocable at will. - Selected evidence: - "When he appointed him over Egypt: the collection of its revenue, fighting its enemy, the reform of its people, and the building of its lands."[134]. - "In people there are faults; the governor is the one most entitled to cover them. Do not uncover what is hidden from you; your duty is to purify what appears; God judges what is hidden from you."[135]. - "And each one, God has named his share and set its limits …"[136]. - "And do not grant to any of your entourage an estate … in a water-right or a common work … whose burden they shift onto others."[137]. - "Do not betray your covenant … and do not let the straitness of an affair in which God's covenant binds you call you to seek its rescission without right."[138]. - Score: 8

10) Amendment meta-rules — Absent - The letter contains no rule specifying who or how foundational norms may be amended. It counsels preserving sound established practices (entrenchment), but sets no amendment procedure. - Score: 0

11) Interpretation and enforcement mechanisms — Present, strong - The letter creates and empowers adjudication

(judges), stands up an interpretive hierarchy (Book and Sunna), requires resources and protection for judges, institutionalizes complaint sessions, and prescribes official oversight and punishment for breaches. - Selected evidence: - "Then choose for judging between people the best among your subjects ..."[139]. - "Then frequently oversee his judgments; give him what removes his need ... and grant him status with you."[140]. - "Refer matters to God and His Messenger ..."[141]. - "Dispatch observers ... if one extends his hand to treachery and your observers' reports concur, punish him in his body ... and brand him with betrayal."[142]. - Score: 9

12) Constitutional conventions (binding political practice) — Present, strong - "Sound sunan" and established ways of governance are treated as binding constraints; the governor is urged to preserve what prior rulers and society found upright. - Selected evidence: - "Do not overturn a sound tradition ... do not introduce a practice that harms past traditions."[143]. - "Increase your study with scholars and converse with sages, to stabilize what your land's affairs have been set right by, and to maintain what people before you found upright."[144]. - "Remember those who preceded you—just government, virtuous practice, and what came down from our Prophet or a prescription in God's Book."[145]. - Score: 8

13) Due process and fair adjudication — Present, strong - The letter requires impartial judges with defined virtues; patience with litigants and scrutiny of evidence; avoidance of precipitous punishment; lawful process in matters of life; and unhindered public access to seek redress. - Selected evidence: - "Then choose for judging ... one whom adversaries do not fluster ... most patient in uncovering matters ... whom praise does not bedazzle nor inducement sway."[146]. - "Do not rush to a sudden [punitive] act when you have an alternative."[147]. - "Beware blood ... there is no excuse ... in willful killing, for it entails retaliation."[148]. - "Sit for them in a public audience ... 'A nation will never be sanctified in which the weak cannot take his right from the strong without stammering.'"[149]. - Score: 8

14) Consent and participation in lawmaking — Partially present (indirect) - The text promotes seeking the public's satisfaction, empowering "the common people," hearing petitions in open audience, and adjusting fiscal burdens upon complaint. But it does not condition enactments or taxes on the formal assent of an assembly or estates; taxes and shares are treated as fixed by the higher law. - Selected evidence: - "Let the most beloved matters to you be those most centered in right, most comprehensive in justice, and most unifying of the subjects' satisfaction—for the public's anger overwhelms the elite's pleasure."[150]. - "The common people are the pillar of religion ... so let your inclination be toward them."[151]. - "If they complain of burdens ... lighten for them."[152]. - Score: 2

15) Stability and continuity across generations — Present, strong - The governor must preserve time-tested norms, anchor decisions in enduring sources (the Book and the Prophet's Sunnah), emulate just predecessors, and cultivate conditions that prolong stable rule. - Selected evidence: - "Do not overturn a sound tradition ..."[153]. - "Increase your study with scholars ... to stabilize what your land's affairs have been set right by ..."[154]. - "Remember those who preceded you—just government, virtuous practice ..."[155]. - Score: 8

16) Abstract commitments enabling adaptation — Present, strong - The charter articulates general moral standards—mercy, justice, equality, fidelity, sanctity of life—whose concrete application can evolve with circumstances. - Selected evidence: - "Imbue your heart with mercy for the subjects ... for they are either your brothers in religion or your peers in creation."[156]. - "Let the most beloved matters to you be those most centered in right and most comprehensive in justice."[157]. - "Do justice to God and to people against yourself."[158]. - "Beware blood and its shedding without right."[159]. - Score: 9

17) Remedies for constitutional breach — Present, substantial - The document prescribes concrete remedies and sanctions for breaches by officials and others: investigative oversight and evidentiary use of informants, bodily punishment, restitution, public shaming for embezzlement; penalties for market manipulation; qisas (retaliation) and delivering the due right to heirs in cases of unlawful killing; and general redress for the oppressed. However, it does not articulate a general doctrine of nullity of ultra vires acts, nor a systematic removal-from-office rule, so the presence is strong but not maximal. - Selected evidence: - "If one of them extends his hand to treachery ... that suffices as a witness; so you inflict bodily punishment upon him ... then set him up in the place of humiliation, and brand him with betrayal."[160]. - "Whoever commits hoarding after your prohibition, then punish him exemplarily, and punish without excess."[161]. - "Do not let the pride of your authority prevent you from giving the slain's guardians their due."[162]. - "For it will be taken from you for others ... and the wronged will be vindicated against you."[163]. - "Whoever wrongs God's servants, God will be his adversary …"[164]. - Score: 7

| Element | Status | Score | Notes |
|---|---|---|---|
| Legal limits on rulers' powers | Present | 9 | Power bound by Book, Sunna, prohibitions on arbitrariness and bloodshed. |
| Sovereignty vs. government offices | Present | 9 | God/higher law above office; governor as delegate. |
| Entrenchment of constraints | Present | 7 | Sacred covenants and preserved sunan; no procedural entrenchment. |
| Writtenness and custom | Present | 9 | Formal "covenant" (ᶜahd); Book and Sunna; established practices. |
| Allocation and checks of power | Present | 8 | Functional differentiation; inspectors; open petitions; judicial independence. |
| Supremacy of higher law | Present | 9 | Disputes referred to Book/Sunna; ruler not above law. |
| Rights against the state | Present | 8 | Bans on oppression, unlawful killing; equality; access to justice; anti-monopoly. |
| Procedural constraints | Present | 9 | Consultation rules, appointment/oversight procedures, public hearings, treaty formalities. |
| Jurisdictional limits | Present | 8 | Defined remit; hidden matters off-limits; common rights protected; covenants binding. |
| Amendment meta-rules | Absent | 0 | No procedure for changing foundational norms. |
| Interpretation and enforcement | Present | 9 | Judges empowered and supported; oversight; grievance sessions; higher-law hierarchy. |
| Constitutional conventions | Present | 8 | Binding force of sound sunan and predecessors' practices. |
| Due process and fair adjudication | Present | 8 | Impartial judges; evidence and patience; no precipitous punishment; qisas. |
| Consent and participation in lawmaking | Partial | 2 | Public satisfaction and petitions encouraged; no formal consent requirement. |
| Stability and continuity | Present | 8 | Preserve tried-and-true norms; emulate just predecessors; rely on enduring sources. |
| Abstract principles enabling adaptation | Present | 9 | General standards (mercy, justice, equality, fidelity, life) guide evolving application. |
| Remedies for breach | Present | 7 | Sanctions for embezzlement/hoarding; qisas; vindication for oppressed; no general nullity rule. |

Table 5: Summary of constitutional elements in Multi-Stage (scores 0-10).

# Notes

[1] ولا تقولن: إني مؤمَّر آمر فأطاع

[2] واردد إلى الله ورسوله ما يضلعك ... فالرد إلى الله: الأخذ بمحكم كتابه، والرد إلى الرسول: الأخذ بسنته ...

[3] فإنهم صنفان: إما أخ لك في الدين، وإما نظير لك في الخلق

[4] ثم اختر لحكمك بين الناس أفضل رعيتك ... ثم أكثر تعاهد قضائه ... وافسح له في البذل ...

[5] واجعل لذوي الحاجات منك قسماً ... وتجلس لهم مجلساً عاماً ... حتى يكلمك متكلمهم غير متعتع

[6] إياك والدماء وسفكها بغير حلها ... ولا عذر لك ... في قتل العمد، لأن فيه قود البدن

[7] فامنع من الاحتكار ... وليكن البيع بيعاً سمحاً: بموازين عدل، وأسعار لا تجحف بالفريقين

[8] فط عهدك بالوفاء، وارع ذمتك بالأمانة ... ولا تغدرن بذمتك

[9] وَارْدُدْ إِلَى اللهِ وَرَسُولِهِ مَا يُضْلِعُكَ

[10] الرَّدُّ إِلَى اللهِ: الأَخْذُ بِمُحْكَمِ كِتَابِهِ

[11] ولا تَقُولَنَّ: إِنِّي مُؤَمَّرٌ آمُرُ فَأُطَاعُ

[12] إمَّا أَخٌ لَكَ في الدِّينِ، وإمَّا نَظِيرٌ لَكَ في الخَلْقِ

[13] وَأَلْزِم الْحَقَّ مَنْ لَزِمَهُ مِنَ الْقَرِيبِ وَالْبَعِيدِ

[14] أَنْصِف... مِنْ نَفْسِك، وَمِنْ خَاصَّةِ أَهْلِكَ

[15] وَمَنْ لَكَ فِيهِ هَوًى مِنْ رَعِيَّتِكَ

[16] اخْتَرْ لِلْحُكْمِ بَيْنَ النَّاسِ أَفْضَلَ رَعِيَّتِكَ

[17] أَكْثِرْ تَعَاهُدَ قَضَائِهِ، وَافْسَحْ لَهُ في الْبَذْلِ

[18] لا تَعْجَلَنَّ إلَى تَصْدِيقِ سَاعٍ

[19] الْوَالِي أَحَقُّ مَنْ سَتَرَهَا

[20] إِيَّاكَ وَالدِّمَاءَ وَسَفْكَهَا بِغَيْرِ حِلِّهَا

[21] فَأَدِّ إلَى أَوْلِيَاءِ الْمَقْتُولِ حَقَّهُمْ

[22] وَاجْعَلْ لِذَوِي الْحَاجَاتِ مِنْكَ قِسْماً

[23] وَتَجْلِسُ لَهُمْ مَجْلِساً عَامّاً

[24] فَلَا تُطَوِّلَنَّ احْتِجَابَكَ عَنْ رَعِيَّتِكَ

[25] الاحْتِجَابُ مِنْهُمْ يَقْطَعُ عَنْهُمْ عِلْمَ مَا احْتَجَبُوا دُونَهُ

[26] وَأَكْثِرْ مُدَارَسَةَ الْعُلَمَاءِ، وَمُنَاقَشَةَ الْحُكَمَاءِ

[27] وَلَا تُدْخِلَنَّ في مَشُورَتِكَ بَخِيلاً

[28] اعْلَمْ أَنَّ الرَّعِيَّةَ طَبَقَاتٌ لا يَصْلُحُ بَعْضُهَا إلَّا بِبَعْضٍ

[29] فَاسْتَعْمِلْهُمُ اخْتِيَاراً

[30] وَلَا تُوَلِّهِمْ مُحَابَاةً وَأُثْرَةً

[31] وَابْعَثِ الْعُيُونَ مِنْ أَهْلِ الصِّدْقِ وَالْوَفَاءِ عَلَيْهِمْ

[32] وَوسَمْتَهُ بِالْخِيَانَةِ

[33] سُخْطُ الْعَامَّةِ يُجْحِفُ بِرِضَا الْخَاصَّةِ

سُخْطُ الخَاصَّةِ يُغْتَفَرُ مَعَ رِضَا العَامَّةِ [34]

اللهَ اللهَ في الطَّبَقَةِ السُّفْلَى [35]

وَاجْعَلْ لَهُمْ قِسْماً مِنْ بَيْتِ مَالِكَ [36]

تَفَقَّدْ أَمْرَ الخَرَاجِ بِمَا يُصْلِحُ أَهْلَهُ [37]

لِيَكُنْ نَظَرُكَ في عِمَارَةِ الأَرْضِ أَبْلَغَ [38]

فَامْنَعْ مِنَ الاحْتِكَارِ [39]

بِمَوَازِينِ عَدْلٍ، وَأَسْعَارٍ لا تُجْحِفُ [40]

وَلا تَقْطَعَنَّ لِأَحَدٍ مِنْ حَاشِيَتِكَ... قَطِيعَةً [41]

وَلا يَطْمَعَنَّ مِنْكَ في اعْتِقَادِ عُقْدَةٍ [42]

لِحَفْظِ عَهْدِكَ بِالوَفَاءِ، وَارْعَ ذِمَّتَكَ بِالأَمَانَةِ [43]

وَلا تَغْدِرَنَّ بِذِمَّتِكَ [44]

لا تَنْقُضْ سُنَّةً صَالِحَةً [45]

وَلا تُحْدِثَنَّ سُنَّةً تَضُرُّ [46]

أَمَرَهُ بِتَقْوَى اللهِ... وَاتِّبَاعِ مَا أَمَرَ بِهِ في كِتَابِهِ: مِنْ فَرَائِضِهِ وَسُنَنِهِ [47]

وَلا تَقُولَنَّ: إِنِّي مُؤَمَّرٌ آمُرُ فَأُطَاعُ [48]

فَإِنَّكَ فَوْقَهُمْ، وَ وَالِي الأَمْرِ عَلَيْكَ فَوْقَكَ، وَاللهُ فَوْقَ مَنْ وَلَّاكَ [49]

أَنْصِفِ اللهَ وَأَنْصِفِ النَّاسَ مِنْ نَفْسِكَ [50]

يَا أَيُّهَا الَّذِينَ آمَنُوا أَطِيعُوا اللهَ وَأَطِيعُوا الرَّسُولَ وَأُولِي الأَمْرِ مِنْكُمْ فَإِنْ تَنَازَعْتُمْ في شَيْءٍ فَرُدُّوهُ إِلَى اللهِ وَالرَّسُولِ [51]

فَإِنَّكَ فَوْقَهُمْ، وَ وَالِي الأَمْرِ عَلَيْكَ فَوْقَكَ، وَاللهُ فَوْقَ مَنْ وَلَّاكَ [52]

وَكُلاًّ قَدْ سَمَّى اللهُ سَبَبَهُ... في كِتَابِهِ أَوْ سُنَّةِ نَبِيِّهِ... عَهْداً مِنْهُ عِنْدَنَا مَحْفُوظاً [53]

وَلا تَنْقُضْ سُنَّةً صَالِحَةً عَمِلَ بِهَا صُدُورُ هذِهِ الأُمَّةِ... وَلا تُحْدِثَنَّ سُنَّةً تَضُرُّ بِشَيْءٍ مِنْ مَاضِي تِلْكَ السُّنَنِ [54]

وَاجْعَلْ نَفْسَكَ جُنَّةً دُونَ مَا أَعْطَيْتَ [55]

فَإِنَّهُ لَيْسَ مِنْ فَرَائِضِ اللهِ عَزَّوَجَلَّ شَيْءٌ... مِنْ تَعْظِيمِ الوَفَاءِ بِالعُهُودِ [56]

هذَا مَا أَمَرَ بِهِ عَبْدُ اللهِ عَلِيٌّ أَمِيرُ المُؤْمِنِينَ، مَالِكَ بْنَ الحَارِثِ الأَشْتَرِ في عهدِهِ إِليه [57]

الرَّدُّ إِلَى اللهِ: الأَخْذُ بِمُحْكَمِ كِتَابِهِ، وَالرَّدُّ إِلَى الرَّسُولِ: الأَخْذُ بِسُنَّتِهِ [58]

وَلا تَنْقُضْ سُنَّةً صَالِحَةً... وَلا تُحْدِثَنَّ سُنَّةً [59]

وَاعْلَمْ أَنَّ الرَّعِيَّةَ طَبَقَاتٌ... مِنْهَا جُنُودُ اللهِ... كُتَّابُ العَامَّةِ وَالخَاصَّةِ... قُضَاةُ العَدْلِ... أَهْلُ الجِزْيَةِ وَالخَرَاجِ... التُّجَّارُ... الطَّبَقَةُ السُّفْلَى [60]

ثُمَّ اخْتَرْ لِحُكْمِ بَيْنَ النَّاسِ أَفْضَلَ رَعِيَّتِكَ [61]

ثُمَّ تَفَقَّدْ أَعْمَالَهُمْ، وَابْعَثِ العُيُونَ مِنْ أَهْلِ الصِّدْقِ وَالوَفَاءِ عَلَيْهِمْ [62]

وَلا تُوَلِّهِمْ مُحَابَاةً وَأَثَرَةً [63]

فَإِنْ تَنَازَعْتُمْ في شَيْءٍ فَرُدُّوهُ إِلَى اللهِ وَالرَّسُولِ [64]

وَاللهُ فَوْقَ مَنْ وَلَّاكَ [65]

وَأَلْزِمِ الْحَقَّ مَنْ لَزِمَهُ مِنَ الْقَرِيبِ وَالْبَعِيدِ[66]

وَلَا تَقُولَنَّ: إِنِّي مُؤَمَّرٌ آمُرُ فَأُطَاعُ[67]

ثُمَّ اللَّهَ اللَّهَ فِي الطَّبَقَةِ السُّفْلَى... وَاجْعَلْ لَهُمْ قِسْمًا مِنْ بَيْتِ مَالِكَ[68]

«لَنْ تُقَدَّسَ أُمَّةٌ لَا يُؤْخَذُ لِلضَّعِيفِ فِيهَا حَقُّهُ مِنَ الْقَوِيِّ غَيْرَ مُتَعْتِعٍ»[69]

إِيَّاكَ وَالدِّمَاءَ وَسَفْكَهَا بِغَيْرِ حِلِّهَا[70]

وَلْيَكُنِ الْبَيْعُ بَيْعًا سَمْحًا، وَأَسْعَارٌ لَا تُجْحِفُ بِالْفَرِيقَيْنِ[71]

وَلَا تَقْطَعَنَّ لِأَحَدٍ مِنْ حَاشِيَتِكَ... قَطِيعَةً، يَكُونُ مَبْدَأُ ذَلِكَ لَهُمْ دُونَكَ، وَعَيْبُهُ عَلَيْكَ[72]

وَاجْعَلْ لِذَوِي الْحَاجَاتِ مِنْكَ قِسْمًا... مَجْلِسًا عَامًّا، تَقْعُدُ عَنْهُمْ جُنْدَكَ وَأَعْوَانَكَ[73]

ثُمَّ انْظُرْ فِي أُمُورِ عُمَّالِكَ، فَاسْتَعْمِلْهُمُ اخْتِبَارًا، وَلَا تُوَلِّهِمْ مُحَابَاةً وَأَثَرَةً[74]

فَإِنْ أَحَدٌ مِنْهُم بَسَطَ يَدَهُ إِلَى خِيَانَةٍ... اكْتَفَيْتَ بِذَلِكَ شَاهِدًا، فَبَسَطْتَ عَلَيْهِ الْعُقُوبَةَ[75]

فَإِنْ تَنَازَعْتُمْ فِي شَيْءٍ فَرُدُّوهُ إِلَى اللَّهِ وَالرَّسُولِ[76]

إِيَّاكَ وَالِاسْتِئْثَارَ بِمَا النَّاسُ فِيهِ أُسْوَةٌ[77]

لَا تُقْطِعَنَّ... قَطِيعَةً... فِي شِرْبٍ أَوْ عَمَلٍ مُشْتَرَكٍ[78]

وَلَا تَدْفَعَنَّ صُلْحًا دَعَاكَ إِلَيْهِ عَدُوُّكَ لِلَّهِ فِيهِ رِضًا[79]

فَامْنَعْ مِنَ الِاحْتِكَارِ، فَإِنَّ رَسُولَ اللَّهِ... مَنَعَ مِنْهُ[80]

وَلَا تَنْقُضْ سُنَّةً صَالِحَةً... وَلَا تُحْدِثَنَّ سُنَّةً تَضُرُّ[81]

ثُمَّ اخْتَرْ لِلْحُكْمِ بَيْنَ النَّاسِ أَفْضَلَ رَعِيَّتِكَ[82]

ثُمَّ أَكْثِرْ تَعَاهُدَ قَضَائِهِ[83]

فَإِنْ تَنَازَعْتُمْ فِي شَيْءٍ فَرُدُّوهُ إِلَى اللَّهِ وَالرَّسُولِ[84]

فَإِنْ أَحَدٌ مِنْهُم بَسَطَ يَدَهُ إِلَى خِيَانَةٍ... فَبَسَطْتَ عَلَيْهِ الْعُقُوبَةَ... وَوَسَمْتَهُ بِالْخِيَانَةِ[85]

وَلَا تَنْقُضْ سُنَّةً صَالِحَةً عَمِلَ بِهَا صُدُورُ هَذِهِ الْأُمَّةِ، وَاجْتَمَعَتْ بِهَا الْأُلْفَةُ[86]

وَالْوَاجِبُ عَلَيْكَ أَنْ تَتَذَكَّرَ مَا مَضَى لِمَنْ تَقَدَّمَكَ: مِنْ حُكُومَةٍ عَادِلَةٍ، أَوْ سُنَّةٍ فَاضِلَةٍ[87]

أَقْلِهِمْ تَبَرُّمًا بِمُرَاجَعَةِ الْخَصْمِ... وَأَصْبِرْهُمْ عَلَى تَكَشُّفِ الْأُمُورِ[88]

وَاجْعَلْ لِذَوِي الْحَاجَاتِ... مَجْلِسًا عَامًّا... حَتَّى يُكَلِّمَكَ مُتَكَلِّمُهُمْ غَيْرَ مُتَعْتِعٍ[89]

وَلَا تُسْرِعَنَّ إِلَى بَادِرَةٍ وَجَدْتَ مِنْهَا مَنْدُوحَةً[90]

فَكُلٌّ بِهِ، وَعَاقِبْ فِي غَيْرِ إِسْرَافٍ[91]

وَلْيَكُنْ أَحَبُّ الْأُمُورِ إِلَيْكَ أَوْسَطُهَا فِي الْحَقِّ... وَأَجْمَعُهَا لِرِضَا الرَّعِيَّةِ[92]

فَإِنَّ سُخْطَ الْعَامَّةِ يُجْحِفُ بِرِضَا الْخَاصَّةِ[93]

وَلَا تَنْقُضْ سُنَّةً صَالِحَةً... وَصَلَحَتْ عَلَيْهَا الرَّعِيَّةُ[94]

وَلَا تَنْقُضْ سُنَّةً صَالِحَةً... وَاجْتَمَعَتْ بِهَا الْأُلْفَةُ، وَصَلَحَتْ عَلَيْهَا الرَّعِيَّةُ[95]

وَالْوَاجِبُ عَلَيْكَ أَنْ تَتَذَكَّرَ مَا مَضَى لِمَنْ تَقَدَّمَكَ: مِنْ حُكُومَةٍ عَادِلَةٍ، أَوْ سُنَّةٍ فَاضِلَةٍ[96]

وَلْيَكُنْ أَحَبُّ الْأُمُورِ إِلَيْكَ أَوْسَطُهَا فِي الْحَقِّ، وَأَعَمُّهَا فِي الْعَدْلِ[97]

وَأَشْعِرْ قَلْبَكَ الرَّحْمَةَ لِلرَّعِيَّةِ [98]

أَنْصِفِ اللهَ وَأَنْصِفِ النَّاسَ مِنْ نَفْسِكَ [99]

فَإِنْ أَحَدٌ مِنْهُمْ بَسَطَ يَدَهُ إِلَى خِيَانَةٍ... اكْتَفَيْتَ بِذَلِكَ شَاهِدًا، فَبَسَطْتَ عَلَيْهِ الْعُقُوبَةَ فِي بَدَنِهِ... ثُمَّ نَصَبْتَهُ بِمَقَامِ الْمَذَلَّةِ، وَوَسَمْتَهُ بِالْخِيَانَةِ [100]

فَمَنْ قَارَفَ حُكْرَةً بَعْدَ نَهْيِكَ إِيَّاهُ فَنَكِّلْ بِهِ، وَعَاقِبْ فِي غَيْرِ إِسْرَافٍ [101]

وَلَا عُذْرَ لَكَ... فِي قَتْلِ الْعَمْدِ، لِأَنَّ فِيهِ قَوَدَ الْبَدَنِ [102]

أَمِرْهُ بِتَقْوَى اللهِ، وَإِيثَارِ طَاعَتِهِ، وَاتِّبَاعِ مَا أَمَرَ بِهِ فِي كِتَابِهِ: مِنْ فَرَائِضِهِ وَسُنَنِهِ [103]

فَإِنَّكَ فَوْقَهُمْ، وَوَالِي الْأَمْرِ عَلَيْكَ فَوْقَكَ، وَاللهُ فَوْقَ مَنْ وَلَّاكَ [104]

وَلَا تَقُولَنَّ: إِنِّي مُؤَمَّرٌ آمُرُ فَأُطَاعُ [105]

إِيَّاكَ وَالدِّمَاءَ وَسَفْكَهَا بِغَيْرِ حِلِّهَا [106]

فَإِنَّكَ فَوْقَهُمْ، وَوَالِي الْأَمْرِ عَلَيْكَ فَوْقَكَ، وَاللهُ فَوْقَ مَنْ وَلَّاكَ [107]

وَارْدُدْ إِلَى اللهِ وَرَسُولِهِ مَا يُضْلِعُكَ مِنَ الْخُطُوبِ... فَالرَّدُّ إِلَى اللهِ: الْأَخْذُ بِمُحْكَمِ كِتَابِهِ، وَالرَّدُّ إِلَى الرَّسُولِ: الْأَخْذُ بِسُنَّتِهِ الْجَامِعَةِ [108]

إِيَّاكَ وَمُسَامَاةَ اللهِ فِي عَظَمَتِهِ، وَالتَّشَبُّهَ بِهِ فِي جَبَرُوتِهِ [109]

وَلَا تَنْقُضْ سُنَّةً صَالِحَةً عَمِلَ بِهَا صُدُورُ هَذِهِ الْأُمَّةِ... وَلَا تُحْدِثْ سُنَّةً تَضُرُّ بِشَيْءٍ مِنْ مَاضِي تِلْكَ السُّنَنِ [110]

وَكُلًّا قَدْ سَمَّى اللهُ سَهْمَهُ، وَوَضَعَ عَلَى حَدِّهِ وَفَرِيضَتِهِ فِي كِتَابِهِ أَوْ سُنَّةِ نَبِيِّهِ... عَهْدًا مِنْهُ عِنْدَنَا مَحْفُوظًا [111]

وَإِنْ عَقَدْتَ بَيْنَكَ وَبَيْنَ عَدُوٍّ لَكَ عُقْدَةً... فَحُطْ عَهْدَكَ بِالْوَفَاءِ... وَاجْعَلْ نَفْسَكَ جُنَّةً دُونَ مَا أَعْطَيْتَ [112]

هَذَا مَا أَمَرَ بِهِ عَبْدُ اللهِ عَلِيٌّ أَمِيرُ الْمُؤْمِنِينَ، مَالِكَ بْنَ الْحَارِثِ الْأَشْتَرِ فِي عَهْدِهِ إِلَيْهِ [113]

وَاتِّبَاعِ مَا أَمَرَ بِهِ فِي كِتَابِهِ: مِنْ فَرَائِضِهِ وَسُنَنِهِ [114]

وَلَا تَنْقُضْ سُنَّةً صَالِحَةً... وَلَا تُحْدِثْ سُنَّةً تَضُرُّ بِشَيْءٍ مِنْ مَاضِي تِلْكَ السُّنَنِ [115]

وَاعْلَمْ أَنَّ الرَّعِيَّةَ طَبَقَاتٌ... فَمِنْهَا جُنُودُ اللهِ، وَمِنْهَا كُتَّابُ الْعَامَّةِ وَالْخَاصَّةِ، وَمِنْهَا قُضَاةُ الْعَدْلِ، وَمِنْهَا عُمَّالُ الْإِنْصَافِ وَالرِّفْقِ [116]

ثُمَّ اخْتَرْ لِلْحُكْمِ بَيْنَ النَّاسِ أَفْضَلَ رَعِيَّتِكَ فِي نَفْسِكَ... [117]

ثُمَّ تَفَقَّدْ أَعْمَالَهُمْ، وَابْعَثِ الْعُيُونَ مِنْ أَهْلِ الصِّدْقِ وَالْوَفَاءِ عَلَيْهِمْ [118]

وَاجْعَلْ لِذَوِي الْحَاجَاتِ مِنْكَ قِسْمًا... وَتَجْلِسُ لَهُمْ مَجْلِسًا عَامًّا... حَتَّى يُكَلِّمَكَ مُتَكَلِّمُهُمْ غَيْرَ مُتَعْتِعٍ [119]

وَارْدُدْ إِلَى اللهِ وَرَسُولِهِ... فَالرَّدُّ إِلَى اللهِ: الْأَخْذُ بِمُحْكَمِ كِتَابِهِ، وَالرَّدُّ إِلَى الرَّسُولِ: الْأَخْذُ بِسُنَّتِهِ الْجَامِعَةِ [120]

فَإِنَّكَ فَوْقَهُمْ، وَوَالِي الْأَمْرِ عَلَيْكَ فَوْقَكَ، وَاللهُ فَوْقَ مَنْ وَلَّاكَ [121]

وَلَا تَنْقُضْ سُنَّةً صَالِحَةً... فَيَكُونَ الْأَجْرُ لِمَنْ سَنَّهَا، وَالْوِزْرُ عَلَيْكَ بِمَا نَقَضْتَ مِنْهَا [122]

وَلَا تَكُونَنَّ عَلَيْهِمْ سَبُعًا ضَارِيًا تَغْتَنِمُ أَكْلَهُمْ [123]

إِيَّاكَ وَالدِّمَاءَ وَسَفْكَهَا بِغَيْرِ حِلِّهَا... وَلَا عُذْرَ لَكَ عِنْدَ اللهِ وَلَا عِنْدِي فِي قَتْلِ الْعَمْدِ، لِأَنَّ فِيهِ قَوَدَ الْبَدَنِ [124]

وَاجْعَلْ لِذَوِي الْحَاجَاتِ مِنْكَ قِسْمًا... «لَنْ تُقَدَّسَ أُمَّةٌ لَا يُؤْخَذُ لِلضَّعِيفِ فِيهَا حَقُّهُ مِنَ الْقَوِيِّ غَيْرَ مُتَعْتِعٍ» [125]

وَامْنَعْ مِنَ الِاحْتِكَارِ... وَلْيَكُنِ الْبَيْعُ بَيْعًا سَمْحًا: بِمَوَازِينِ عَدْلٍ، وَأَسْعَارٍ لَا تَجْحَفُ بِالْفَرِيقَيْنِ [126]

وَأَلْزِمِ الْحَقَّ مَنْ لَزِمَهُ مِنَ الْقَرِيبِ وَالْبَعِيدِ [127]

وَلَا تَدْخُلَنَّ فِي مَشُورَتِكَ بَخِيلًا... وَلَا جَبَانًا... وَلَا حَرِيصًا [128]

وَلَا تَعْجَلَنَّ إِلَى تَصْدِيقِ سَاعٍ [129]

وَاجْعَلْ لِذَوِي الْحَاجَاتِ مِنْكَ قِسْمًا... وَتَجْلِسُ لَهُمْ مَجْلِسًا عَامًّا... حَتَّى يُكَلِّمَكَ مُتَكَلِّمُهُمْ غَيْرَ مُتَعْتِعٍ [130]

فَنْ قَارَفَ حُكْرَةً بَعْدَ نَهْيِكَ إِيَّاهُ فَنَكِّلْ بِهِ، وَعَاقِبْ فِي غَيْرِ إِسْرَافٍ [131]

وَلَا تَعْقِدْ عَقْدًا تَجَوِّزُ فِيهِ الْعِلَلَ، وَلَا تَعَوِّلَنَّ عَلَى لَحْنِ الْقَوْلِ بَعْدَ التَّأْكِيدِ وَالتَّوْثِقَةِ [132]

فَإِنْ شَكَوْا ثِقَلًا أَوْ عِلَّةً... خَفَّفْتَ عَنْهُمْ بِمَا تَرْجُو أَنْ يَصْلُحَ بِهِ أَمْرُهُمْ [133]

حين ولاّه مصر: جبوة خراجها، وجهاد عدوّها، واستصلاح أهلها، وعمارة بلادها[134]

فإنّ في الناس عيوباً، الوالي أحقّ من سترها عمّا غاب عنك منها، فلا تكشفنّ عمّا غاب عنك من سترها، فإنما عليك تطهير ما ظهر لك، والله يحكم على ما غاب عنك[135]

وكلاً قد سمّى الله سهمه، ووضع على حدّه وفريضته...[136]

ولا تقطعنّ لأحد من حاشيتك وحامتك قطيعة... في شِرب أو عمل مشترك، يحملون مؤونته على غيرهم[137]

فلا تغدرنّ بذمّتك... ولا يدعونّك ضيق أمر لزمك في طلب انفصاحه بغير الحق[138]

ثم اختر للحكم بين الناس أفضل رعيتك في نفسك...[139]

ثم أكثر تعاهد قضائه، وافسح له في البذل ما يزيل علّته... وأعطه من المنزلة لديك...[140]

واردد إلى الله ورسوله...[141]

وابعث العيون... فإن أحداً منهم بسط يده إلى خيانة... فبسطت عليه العقوبة في بدنه... ووسمته بالخيانة[142]

ولا تنقض سنة صالحة... ولا تحدثنّ سنة تضرّ بشيء من ماضي تلك السنن[143]

وأكثر مدارسة العلماء، ومنافثة الحكماء، في تثبيت ما صلح عليه أمر بلادك، وإقامة ما استقام به الناس قبلك[144]

والواجب عليك أن تتذكّر ما مضى لمن تقدّمك: من حكومة عادلة، أو سنة فاضلة، أو أثر عن نبينا... أو فريضة في كتاب الله[145]

ثم اختر للحكم بين الناس أفضل رعيتك... وأصبرهم على تكشّف الأمور... ممن لا يزدهيه إطراء، ولا يستميله إغراء[146]

ولا تسرعنّ إلى بادرة وجدتَ منها مندوحة[147]

إياك والدماء... ولا عذر لك... في قتل العمد، لأن فيه قود البدن[148]

وتجلس لهم مجلساً عاماً... «لن تقدّس أمة لا يؤخذ للضعيف فيها حقّه من القوي غير متعتع»[149]

ولكن أحب الأمور إليك أوسطها في الحق، وأعمها في العدل، وأجمعها لرضا الرعية، فإن تخط العامة يجحف برضا الخاصة[150]

وإنما عمود الدين... العامة من الأمة، فليكن صغوك لهم، وميلك معهم[151]

فإن شكوا ثقلاً... خففت عنهم[152]

ولا تنقض سنة صالحة...[153]

وأكثر مدارسة العلماء... في تثبيت ما صلح عليه أمر بلادك...[154]

والواجب عليك أن تتذكّر ما مضى لمن تقدّمك: من حكومة عادلة، أو سنة فاضلة...[155]

وأشعر قلبك الرحمة للرعية... فإنهم صنفان: إما أخ لك في الدين، وإما نظير لك في الخلق[156]

ولكن أحب الأمور إليك أوسطها في الحق، وأعمها في العدل[157]

أنصف الله وأنصف الناس من نفسك[158]

إياك والدماء وسفكها بغير حلها[159]

فإن أحداً منهم بسط يده إلى خيانة... اكتفيتَ بذلك شاهداً، فبسطت عليه العقوبة في بدنه... ثم نصبته بمقام المذلة، ووسمته بالخيانة[160]

فن قارف حكرة بعد نهيك إياه فنكّل به، وعاقب في غير إسراف[161]

فلا تطمحنّ بك نخوة سلطانك عن أن تؤدّي إلى أولياء المقتول حقهم[162]

فإنه مأخوذ منك لغيرك... وينتصف منك للمظلوم[163]

ومن ظلم عباد الله كان الله خصمه دون عباده...[164]